\documentclass[journal]{IEEEtran}
\usepackage{blindtext}
\usepackage{amsmath, psfrag}
\usepackage{amssymb, relsize}
\usepackage{textcomp}
\usepackage{graphicx}
\usepackage[export]{adjustbox}
\usepackage{subcaption,dirtytalk}
\usepackage{tikz}
\usepackage{pgfplots}
\usepackage{setspace}
\usepackage{multirow}
\usepackage{epstopdf,caption}
\newcommand{\subparagraph}{}
\usepackage{titlesec}
\usepackage{etoolbox}
\usepackage{multirow}
\usepackage{pgf,tikz,pgfplots}
\usepackage{mathrsfs}
\usetikzlibrary{arrows}
\usetikzlibrary{positioning}
\usetikzlibrary{shapes.geometric, arrows.meta, decorations.markings}
\tikzstyle{startstop} = [rectangle, rounded corners, minimum width=1.25cm, minimum height=0.75cm,text centered, draw=black, fill=red!20, node distance = 1cm]
\tikzstyle{startstop1} = [rectangle, rounded corners, minimum width=1cm, minimum height=0.5cm,text centered, draw=black, fill=blue!30, node distance = 1cm]
\tikzstyle{startstop2} = [rectangle, rounded corners, minimum width=10.5cm, minimum height=1.5cm,text centered, draw=black, fill=gray!10, node distance = 1.5cm]
\tikzstyle{arrow} = [thick,->,>=stealth]
\makeatletter
\patchcmd{\ttlh@hang}{\parindent\z@}{\parindent\z@\leavevmode}{}{}
\patchcmd{\ttlh@hang}{\noindent}{}{}{}
\makeatother
\ifCLASSINFOpdf
 
\else
  
\fi

\begin{document}
\definecolor{dbwrru}{rgb}{0.8588235294117647,0.3803921568627451,0.0784313725490196}
\definecolor{rvwvcq}{rgb}{0.08235294117647059,0.396078431372549,0.7529411764705882}
\author{Neema Davis, Gaurav Raina, Krishna Jagannathan \thanks{N. Davis, G. Raina and K. Jagannathan are with the Department
of Electrical Engineering, Indian Institute of Technology Madras, Chennai 600 036, India. 
E-mail: \{ee14d212, gaurav, krishnaj\}@ee.iitm.ac.in}}
\title{Grids versus Graphs: Partitioning Space for Improved Taxi Demand-Supply Forecasts }

\maketitle
\begin{abstract}
Accurate taxi demand-supply forecasting is a challenging application of ITS (Intelligent Transportation Systems), due to the complex spatial and temporal patterns. 
We investigate the impact of different spatial partitioning techniques on the prediction performance of an LSTM (Long Short-Term Memory) network, in the context of taxi demand-supply forecasting.
We consider two tessellation schemes: (i) the variable-sized Voronoi tessellation, and (ii) the fixed-sized Geohash tessellation.  
While the widely employed ConvLSTM (Convolutional LSTM) can model fixed-sized Geohash partitions, the standard convolutional filters cannot be applied on the variable-sized Voronoi partitions. 
To explore the Voronoi tessellation scheme, we propose the use of GraphLSTM (Graph-based LSTM), by representing the Voronoi spatial partitions as nodes on an arbitrarily structured graph. 
The GraphLSTM offers competitive performance against ConvLSTM, at lower computational complexity, across three real-world large-scale taxi demand-supply data sets, with different performance metrics. 
To ensure superior performance across diverse settings, a HEDGE based ensemble learning algorithm is applied over the ConvLSTM and the GraphLSTM networks.


\end{abstract}


\IEEEpeerreviewmaketitle

\section{Introduction} \label{intro}
\begin{figure}[ht!]
\centering
        \psfrag{t}{\hspace{-10mm} \raisebox{-2mm}{\footnotesize{Time of the day}}}
        \psfrag{v}{\hspace{-6mm}\raisebox{1mm}{\footnotesize{Magnitude}}}
        \psfrag{S}{\hspace{0mm}\raisebox{-0.4mm}{\scalebox{0.8}{Supply}}}
        \psfrag{D}{\hspace{0mm}\raisebox{-0.5mm}{\scalebox{0.8}{Demand}}}
            	\psfrag{1}{\hspace{0mm}\raisebox{-0.7mm}{\footnotesize{1}}}
            	 \psfrag{12}{\hspace{0mm}\raisebox{-0.7mm}{\footnotesize{12}}}
            	\psfrag{24}{\hspace{0mm}\raisebox{-0.7mm}{\footnotesize{24}}}
            	\psfrag{25}{\hspace{-2mm}\raisebox{-0.1mm}{\footnotesize{25}}}
            	\psfrag{49}{\hspace{-2mm}\raisebox{-0.1mm}{\footnotesize{50}}}
            	\psfrag{280}{\hspace{-2mm}\raisebox{-0.1mm}{\footnotesize{280}}}
        \includegraphics[height = 1.5in, width = 2.25in]{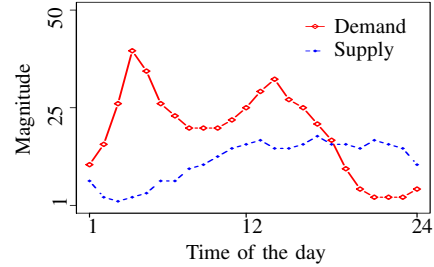}
    \caption{The demand-supply patterns near Bengaluru city center averaged over all Mondays, where the mismatch between the demand for taxis and the available supply is visible. Our study is motivated by this demand-supply imbalance, which can be mitigated by accurate demand-supply forecasts.}
    \label{sdmismatch}
 \end{figure}

Spatio-temporal forecasting has a wide range of applications, ranging from epidemic detection \cite{colborn2018spatio}, energy management \cite{ezzat2018spatio}, to cellular traffic \cite{wang2018spatio}, among others. Location-based taxi demand and supply forecasting, one of the key components of ITS (Intelligent Transportation Systems), also relies heavily on accurate spatio-temporal forecasting. Mobility-on-Demand services such as e-hailing taxis, which have gained tremendous popularity in the recent years, often face taxi demand-supply imbalances.  During peak and off-peak hours, mismatches occur between the spatial distributions of the taxi demand and the available drivers, resulting in either scarcity or abundance of vacant taxis. For example, Fig. \ref{sdmismatch} presents a case of demand-supply mismatch averaged over all Mondays near the city center in Bengaluru, India.  We see that during the day hours, the high demand for taxis is met with inadequate supply. On the other hand, there is a surplus of vacant taxis during night hours, against low customer demand. Accurate demand-supply forecasts can mitigate this imbalance, thereby improving the efficiency of these taxi services. Information regarding the expected future demand and supply in a region can be used to re-route vacant cruising taxis, dynamically adjust the taxi fares, and recommend popular pick-up locations to the drivers.

\subsection{Related works}
The recent popularity of e-hailing taxi services has generated substantial interest in developing efficient taxi demand-supply prediction algorithms \cite{kamga2015analysis, jager2016analyzing, tong2017simpler, ke2017short}. In the past, taxi demand-supply prediction was mainly formulated as a classical time-series forecasting problem. ARMA (Auto Regressive and Moving Average) family of time-series models were applied to taxi demand prediction problems with satisfactory results \cite{nagy2018survey}. However, these time-series models rely on the stationarity assumption, which is often violated by real-world data. The capability of such classical methods to deal with high dimensional, complex, and dynamic time-series data is also limited. Meanwhile, the generalization capabilities of the NNs (Neural Networks) inspired transportation researchers to leverage this tool in the traffic forecasting domain with promising results \cite{nguyen2018deep, wang2018enhancing}. The traditional NNs lack the ability to learn temporal dependencies, leading to the design of models that are more suited for sequence data such as RNN (Recurrent Neural Network) and its variants, namely LSTM (Long Short-Term Memory) and GRU (Gated Recurrent Unit) \cite{petnehazi2019recurrent, fu2016using}. In fact, RNN has emerged as the preferred machine learning tool to solve many traditional sequence learning problems such as speech recognition \cite{chiu2018state}, text recognition \cite{yogatama2017generative} and cellular communication \cite{rutagemwa2018dynamic}. Since the real-world data often exhibit both temporal and spatial variations, several spatio-temporal extensions of RNNs have been proposed. A widely employed extension used in taxi data forecasting, known as the ConvLSTM (Convolutional LSTM), involves addition of convolutional layers prior to the LSTM framework \cite{xingjian2015convolutional}. The standard convolutional layer can be applied to only a grid-structured input and learns localized rectangular filters. This limits the application of the conventional ConvLSTM to a city space partitioned into fixed-sized grids. Hence, a fixed-sized equally-spaced partitioning is often adopted in the spatio-temporal NN-based models for location-based taxi demand or supply forecasts \cite{yao2018deep, liao2018large, ke2017short, wang2017deepsd, wang2018deepstcl}.

In our previous work \cite{davis2018ataxi}, we explored a variable-sized partitioning scheme in addition to a fixed-sized scheme for taxi demand forecasting. Using classical time-series regression models, we observed a visible enhancement in the prediction performance with a variable-sized tessellation scheme in several scenarios. The real-world data often has a heterogeneous spatial distribution, which may not be captured faithfully with a fixed-sized partitioning scheme that is based on spatial homogeneity assumption. While the generalization capabilities of the RNNs make them powerful tools for spatio-temporal modeling, assuming that the data is homogeneously distributed may limit their modeling capabilities. Hence, it is imperative to explore a variable-sized partitioning scheme in an RNN-based spatio-temporal modeling framework. Most of the currently popular spatio-temporal RNN models are based on the ConvLSTM networks that are incapable of modeling a variable-sized partitioned space. Motivated by this observation, in this work, we develop an LSTM framework that can extract the potential of variable-sized spatial partitions. 

While dividing the city space into variable-sized Voronoi tessellations, we take note of the fact that arbitrarily spaced tessellations can be represented using graphs. That is, while the variable-sized partitions cannot be represented as equally-spaced fixed-sized grids, they can be visualized in the form of a graph. The demand aggregated in each Voronoi partition can form a node in an arbitrary structured graph. Therefore, a Graph-based RNN holds great potential in our scenario. In the last couple of years, there has been substantial interest in devising Graph NNs, by extending the convolution operator to suit a more general graph-structured data \cite{wu2019comprehensive}. In the context of traffic forecasting, Graph CNNs (Convolutional Neural Networks) have been applied to predict flows at traffic sensors \cite{li2018diffusion}. By considering a road network as a graph and traffic sensors as nodes, Graph CNNs have been combined with RNN to capture the spatial relationships between nodes \cite{li2018diffusion, cui2018high, wang2018dynamic}. However, there is limited research on incorporating graph RNNs in location-based taxi demand or supply forecasting. In \cite{geng2019spatiotemporal}, the authors do apply graphs to model non-euclidean correlations for ride-hailing demand forecasting, but the models are based on equally-spaced fixed-sized grid partitions. In summary, the existing transportation literature on Graph NNs either consider traffic sensors as the graph nodes or learn graph-based correlations in a grid-partitioned space. 

Our modeling framework deviates significantly from the existing literature as we employ a GraphLSTM (Graph-based LSTM) \cite{cui2018high} model to learn an arbitrarily structured graph, where each node corresponds to the aggregated demand in a spatial Voronoi partition. To the best of our knowledge, in the context of spatial partitioning, Graph-based RNNs have not been explored in the literature. Another important contribution of this paper lies in understanding the impact of different spatial partitioning schemes on the predictive performance of RNNs. To that end, we perform a comparison of the Geohash-based ConvLSTM and the Voronoi-based GraphLSTM. These features set our work apart from the existing literature. \footnote{A part of this work was presented as a poster at the NIPS Workshop on Machine Learning in Intelligent Transportation Systems, 2018 \cite{davis2018btaxi}.}

\subsection{Our contributions}\label{contributions}
After the city is divided into fixed-sized rectangular cells and variable-sized polygon cells, we employ the standard ConvLSTM to model the equally-spaced Geohash tessellated city and the GraphLSTM to model the unequally-spaced Voronoi tessellated city. We compare the results with three baselines: (i) the \emph{vanilla} LSTM based on Voronoi and Geohash schemes, (ii) the ARIMA (Auto Regressive Integrated Moving Average) model, and (iii) the ARIMAX (ARIMA with eXogenous inputs) model. When evaluated across three real-world data sets, the GraphLSTM exhibits competitive prediction performance against the established baseline models, at a lower computational complexity. Interestingly, we see that the prediction models exhibit non-stationary behavior, in addition to dependencies on the choice of data set and performance metric. To tackle this issue, we perform ensemble learning on the time-shifting models using an online non-stationary expert combining dHEDGE algorithm \cite{raj2017aggregating}. By using a combination of prediction models, the algorithm picks the best model for each time step in the forecasting horizon. The main contributions of this paper are the following:
\begin{itemize}
    \item This work is the first to demonstrate the potential of Graph RNNs within a location-based spatial partitioning and forecasting framework.
    \item The GraphLSTM offers competitive prediction performance against ConvLSTM at a lower computational complexity, across data sets using different performance metrics.
    \item The Voronoi-based GraphLSTM outperforms Geohash-based GraphLSTM and ConvLSTM in data scarce locations.
    \item Prediction accuracy of irregular graph based GraphLSTM is at least as good as that of regular graph based GraphLSTM, highlighting the potential of irregular graphs in location-based forecasting.
    \item Applying the dHEDGE algorithm in conjunction with the ConvLSTM and GraphLSTM models ensure consistently superior prediction accuracy, across all scenarios considered.
\end{itemize}

The rest of the paper is organized as follows. Section \ref{problemsetting} defines the problem statement. The spatial tessellation schemes are explained in Section \ref{tessellation}, along with a brief description of the data sets used in this study. In Section \ref{modeling}, the spatio-temporal LSTM, ConvLSTM, GraphLSTM  and the baseline models are discussed. The experimental settings and results are elaborated in Section \ref{experiments}, followed by a description of the dHEDGE algorithm in Section \ref{dhedge}. We conclude our results in Section \ref{conclusions}. 
\begin{figure}[ht!]
\centering
\scalebox{0.9}{
\begin{tikzpicture}[every text node part/.style={align=center},decoration={markings,mark=at position 1 with {\arrow[scale=2,black]{latex}};}]
   \node[startstop] (title) {Spatio-Temporal Data};
    \node[startstop,below left = of title, xshift = 1.5cm] (V) {Variable-sized Voronoi \\ Tessellation};
    \node[startstop,below right = of title, xshift = -1.75cm] (G) {Fixed-sized Geohash \\ Tessellation};
    \node [startstop, below = of V] (graph){Spatio-Temporal \\ GraphLSTM Model};
    \node [startstop, below = of G] (grid){Spatio-Temporal \\ ConvLSTM Model};
    \node [startstop, below right = of graph, yshift = -0cm, xshift = -2cm] (baselines){dHEDGE Combining Algorithm};

    \draw[postaction={decorate}] (title.south) -- ++(0,0) -- ++(0,-0.3) -|  (V.north);
    \draw[postaction={decorate}] (title.south) -- ++(0,0) -- ++(0,-0.3) -|  (G.north);
    \draw (V)       edge[postaction={decorate}] (graph);
    \draw (G)       edge[postaction={decorate}] (grid);
    \draw[postaction={decorate}] (graph.south) -- ++(0,0) -- ++(0,-0.3) -|  (baselines.north);
    \draw[postaction={decorate}] (grid.south) -- ++(0,0) -- ++(0,-0.37) -|  (baselines.north);
\end{tikzpicture}}
\caption{Flow chart of the study in this paper. The spatio-temporal data is first tessellated using the Voronoi and Geohash schemes. After modeling and comparing the tessellated data using convolutional and graphical LSTM models, ensemble learning is performed, which helps to achieve superior prediction performance. }
\label{flowchart}
\end{figure}
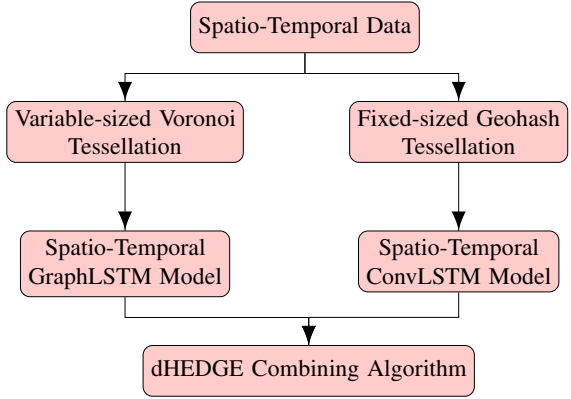

\section{Problem setting} \label{problemsetting}
We formulate our problem as follows. For a location-based forecasting, the city space is tessellated into $N$ regions. The set of regions $R$ = $\{r_1,...,r_N\}$ can be fixed-sized grids, variable-sized polygons, zip codes, etc. We employ (i) fixed-sized rectangular grids called \emph{geohashes} and (ii) variable-sized polygon partitions called \emph{Voronoi cells}. Let the demand and supply aggregated in every $r_i \in R$ form the sets $D$ = $\{d_1,...,d_N\}$ and $S$ = $\{s_1,...,s_N\}$. We assume that the data in the region of interest is related to its historical data and the data in its first-order neighboring regions. In this work, our objectives are two-fold. First, given the set of all geohashes, our goal is to learn a function $\mathcal{F(\cdot)}$, mapping the demand (or supply) data in any geohash to its temporal and spatial neighbors. For the Geohash-based fixed-sized equally-spaced spatial structure, we use ConvLSTM to learn this function $\mathcal{F(\cdot)}$ as:
\begin{equation*}
    \mathcal{F}([d_{1:t,1:N}]|geohashes) = [d_{t+1:t+h, 1:N}],
\end{equation*}
where, $h$ is the forecast horizon. Second, for exploring Voronoi-based variable-sized unequally-spaced spatial structure, we represent the tessellated city space by an undirected graph $\mathcal{G}$, where $\mathcal{G}$ = ($\mathcal{V}$, $\mathcal{E}$, $\mathcal{A}$). The $N$ regions will form a graph with $\mathcal{V}$ vertices and $\mathcal{E}$ edges. The connectivity between nodes is represented by an adjacency matrix $\mathcal{A} \in \mathbb{R}^{N\times N}$. The adjacency matrix is defined as follows:
\begin{equation*}
    \mathcal{A}_{i,j} = \begin{cases}
      1 & \text{if there is a link connecting nodes $i$ and $j$},\\
    0 & \text{otherwise.}
    \end{cases}
\end{equation*}
By default, $A_{i,i} = 0$. For the graph learning task, our choice of modeling tool is the GraphLSTM. Here, the taxi demand-supply forecasting problem can be represented as learning the mapping function $\mathcal{F(\cdot)}$ that maps the historical demand (or supply) to future predictions, given a graph depicting the Voronoi partitions:
\begin{equation*}
    \mathcal{F}([d_{1:t,1:N}]|\mathcal{G}(\mathcal{V}, \mathcal{E}, \mathcal{A})) = [d_{t+1:t+h, 1:N}].
\end{equation*}
The predictive performance of the models is compared using three error metrics, namely Symmetric Mean Absolute Percentage Error (SMAPE), Mean Absolute Scaled Error (MASE) and Root Mean Square Error (RMSE). These metrics are defined and discussed in Section \ref{modeling}. Fig. \ref{flowchart} shows the flow chart of the study to be conducted in this paper.

\section{Spatial Tessellation schemes}\label{tessellation}
Three real-world data sets are considered for our study. We use the taxi demand-supply data sets from the city of Bengaluru, India and publicly available taxi demand data set from the city of New York, USA. 

\subsection{Description of the data sets}
The Bengaluru taxi demand and driver supply data sets are acquired from a leading Indian e-hailing taxi service provider. The demand data contains GPS traces of taxi passengers booking a taxi by logging into their mobile application. The supply data contains GPS traces of fresh log-ins of taxi drivers, representing available supply. The data sets are available for a period of two months; from $\text{1}^{st}$ of January 2016 to $\text{29}^{th}$ of February 2016. The data sets contain latitude-longitude coordinates of the passenger/driver, session duration and time stamp. The latitude and longitude coordinates of the city are 12.9716° N, 77.5946° E, with an area of approximately 740 $\text{km}^2$. The publicly available New York yellow taxi data set \cite{nyc} contains GPS traces of a street hailing yellow taxi service.  For our study, we extract the pick-up locations and time stamps from the period of January-February 2016.  The latitude and longitude coordinates of the New York city are 40.7128° N, 74.0059° W, with an area of approximately 780 $\text{km}^2$. Some key statistical properties of the data sets are given in Table \ref{charac}.
\begin{table}[t!]
\scalebox{0.85}{
{\renewcommand{\arraystretch}{1.1}
\begin{tabular}{|l||c|c|c|}
\hline
\begin{tabular}[c]{@{}c@{}} Statistical Properties \end{tabular}  & \begin{tabular}[c]{@{}c@{}}Bengaluru\\ Demand \end{tabular}  & \begin{tabular}[c]{@{}c@{}}Bengaluru\\ Supply \end{tabular}  & \begin{tabular}[c]{@{}c@{}}New York\\ Demand \end{tabular} \\ \hline \hline
Minimum         & 0       & 0       &  0   \\ \hline
Maximum         & 630       &  2913      &  1582   \\ \hline
Mean        &  14.1      &   10.1     &  19.5   \\ \hline
Median        &  13      & 7       &  16   \\ \hline
Skewness        & 0.77       &  5.78      &   1.08   \\ \hline
Kurtosis        &  1.31       &  114.5      & 3.56    \\ \hline
Standard Deviation & 10.1      &   9.07        & 14.9    \\ \hline
Periodicity (in time steps) & 12, 24       &  12, 24, 168      & 12, 24, 168    \\ \hline
\end{tabular}}}
\caption{Statistical properties of the taxi demand-supply data sets from Bengaluru and New York City, when aggregated as time-sequences over 60 minutes for 60 days. }
\label{charac}
\end{table}
\begin{figure*}[htb]
\begin{subfigure}{0.385\textwidth}
\begin{tikzpicture}
\begin{axis}[enlargelimits=false, 
                 axis on top,
                 height = 4.5cm,
                 width=0.9\textwidth,
                 ylabel={\footnotesize{Longitude}},
                 xlabel={\footnotesize{Latitude}},
                  ylabel near ticks,
                xlabel near ticks,
                 ytick={0.1,0.9},
                 xticklabels={\scriptsize{12.75},\scriptsize{13.15}},
                 xtick={76.1,76.9},
                 yticklabels={\scriptsize{77.45},\scriptsize{77.8}},
                 yticklabel style={rotate=0}
                  ]
    \addplot graphics [ymin=0,ymax=1,xmin=76,xmax=77]{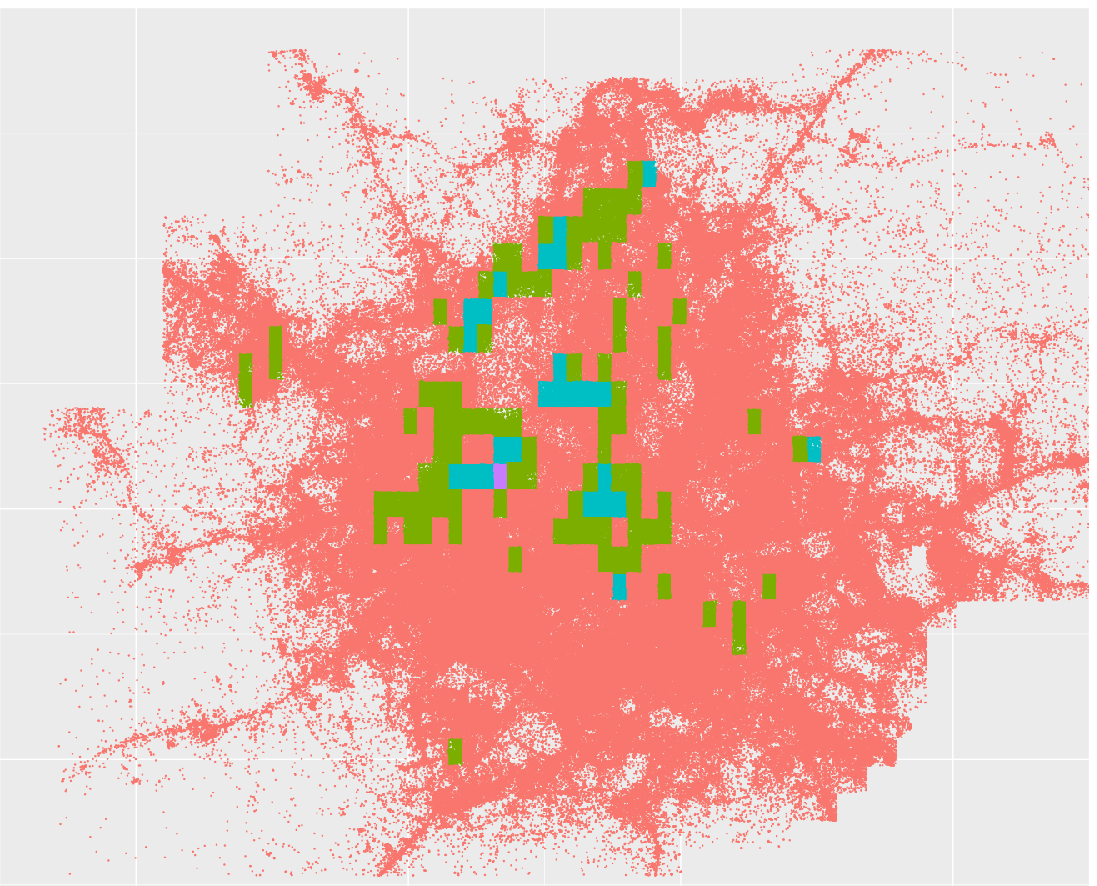};
    \end{axis}
 \end{tikzpicture}
 \vspace{-0.2cm}
  \caption{Geohash tessellation}
\label{voronoi_hm}
\end{subfigure}%
\begin{subfigure}{0.385\textwidth}
\begin{tikzpicture}
\begin{axis}[enlargelimits=false, 
                 axis on top,
                 height = 4.5cm,
                 width=0.9\textwidth,
                 ylabel={\footnotesize{Longitude}},
                 xlabel={\footnotesize{Latitude}},
                  ylabel near ticks,
                xlabel near ticks,
                 ytick={0.1,0.9},
                 xticklabels={\scriptsize{12.75},\scriptsize{13.15}},
                 xtick={76.1,76.9},
                 yticklabels={\scriptsize{77.45},\scriptsize{77.8}},
                 yticklabel style={rotate=0}
                  ]
    \addplot graphics [ymin=0,ymax=1,xmin=76,xmax=77]{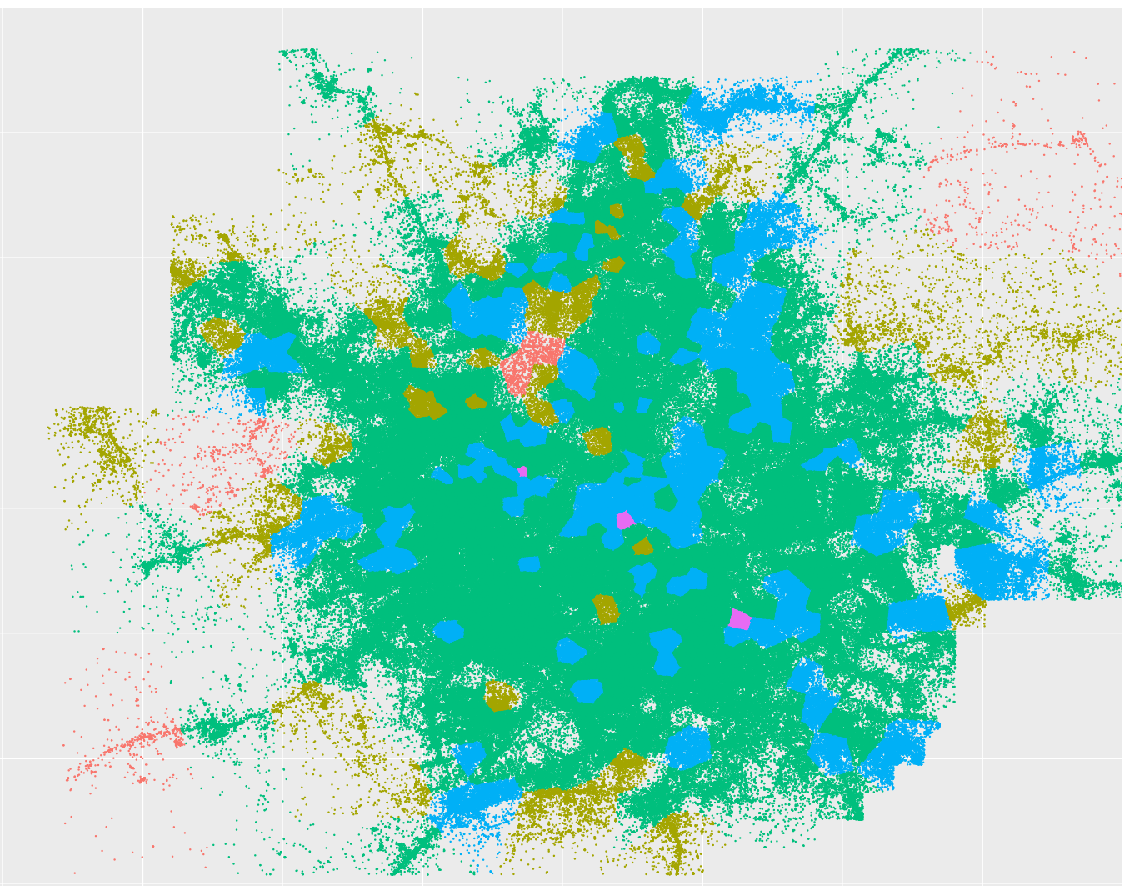};
    \end{axis}
\end{tikzpicture}
\vspace{-0.2cm}
 \caption{Voronoi tessellation}
\label{geohash_hm}
\end{subfigure}%
\begin{subfigure}{0.2\textwidth}
\begin{picture}(100,100)
  \put(15,30)  {\includegraphics[scale=0.175]{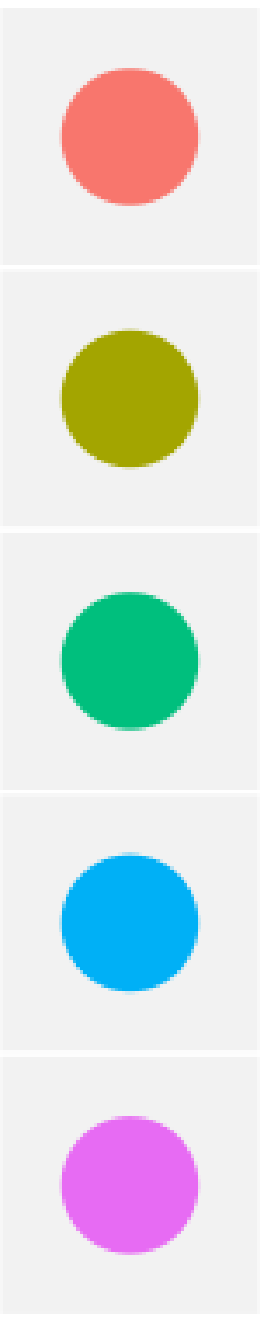}}
  \put(35,87) {\footnotesize{0-5000}}
  \put(35,75) {\footnotesize{5000-10000}}
  \put(35,60) {\footnotesize{10000-25000}}
  \put(35,47) {\footnotesize{25000-50000}}
  \put(35,32) {\footnotesize{50000-150000}}
\end{picture}
\caption{Data per region}
\end{subfigure}%
\vspace{3mm}
\caption{Heat maps obtained by partitioning the city of Bengaluru into Voronoi cells and 6-level geohashes. The spatial distribution of data in the partitions varies significantly with the tessellation scheme employed. While the Geohash scheme produces several data dense and scarce partitions, the Voronoi scheme uniformly distributes data in each partition.} \label{heatmaps}
\end{figure*}

\subsection{Voronoi tessellation}
A prerequisite for the Voronoi tessellation is a set of generating sites that can be used to define the Voronoi cells. For this purpose, we use the K-Means clustering algorithm \cite{macqueen1967some}. The algorithm has a linear memory and time complexity, which is ideal for our very large data sets, and performs reasonably well in comparison with other clustering algorithms \cite{chaudhari2012comparative}. The data is classified into a predefined set of clusters, and the centroids of these clusters act as generating sites for the Voronoi tessellation. 

The K-Means algorithm aims to minimize the squared error function $J$ given as:
\begin{equation}
    J = \sum_{j=1}^k \sum_{i=1}^n ||y_{(i)}^j - c_j||^2,
\end{equation}
where $||y_{(i)}^j - c_j||^2$ is the Euclidean distance between a data point $y_{(i)}^j$ and its center $c_j$, and $n$ is the total number of data points. For efficient rerouting of vacant taxi drivers, we partition Bengaluru into 740 clusters and New York city into 780 clusters so that the average cluster area remains close to 1  $\text{km}^2$. Note that instead of applying a separate K-Means algorithm on the Bengaluru supply data set, we associate the supply data points with its nearest demand centroid. It enables us to do a comparative analysis of the demand and supply patterns associated with every demand region of interest. Then, the Centroidal Voronoi tessellation divides the space according to the nearest neighbor-rule, based on the K-Means centroids. Based on the closeness of centroids, this tessellation strategy produces polygon partitions of varying areas, with a time complexity of O(nlogn). 

\subsection{Geohash tessellation}
Geohash tessellation is an extension of the basic grid partitioning technique with a naming convention. Each latitude-longitude coordinate is encoded into an alphanumeric string, where the length of the string denotes the \emph{level} of the geohash. All latitude and longitude coordinates mapped to a specific string will form a unique fixed-sized rectangular grid. For example, a 5-level geohash spans an area of 4.9 km $\times$ 4.9 km and a 6-level geohash covers an area of 1.2 km $\times$ 0.6 km. For consistent comparison with Voronoi cells of average area 1 $\text{km}^2$, we employ 6 level-geohashes for our study. Regarding time complexity, this algorithm is O(1). 

The Voronoi and Geohash heat maps are plotted in Fig. \ref{heatmaps}, where the scale denotes the data volume in each cell. Geohash strategy produces rectangular grids of fixed area, resulting in several demand dense and scarce cells. Voronoi strategy tends to uniformly distribute data and produces polygons of variable area. 
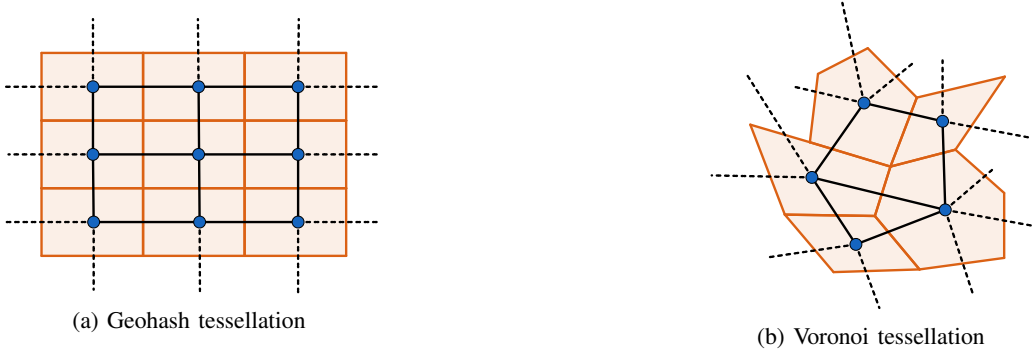
\begin{figure*}[htb]
\begin{subfigure}{0.5\textwidth}
\centering
\scalebox{0.45}{
\begin{tikzpicture}[line cap=round,line join=round,>=triangle 45,x=1cm,y=1cm]
\fill[line width=2pt,color=dbwrru,fill=dbwrru,fill opacity=0.1] (-4,2) -- (-4,0) -- (-1,0) -- (-1,2) -- cycle;
\fill[line width=2pt,color=dbwrru,fill=dbwrru,fill opacity=0.1] (-1,2) -- (2,2) -- (2,0) -- (-1,0) -- cycle;
\fill[line width=2pt,color=dbwrru,fill=dbwrru,fill opacity=0.1] (2,2) -- (5,2) -- (5,0) -- (2,0) -- cycle;
\fill[line width=2pt,color=dbwrru,fill=dbwrru,fill opacity=0.1] (-4,0) -- (-4,-2) -- (-1,-2) -- (-1,0) -- cycle;
\fill[line width=2pt,color=dbwrru,fill=dbwrru,fill opacity=0.1] (-1,-2) -- (2,-2) -- (2,0) -- (-1,0) -- cycle;
\fill[line width=2pt,color=dbwrru,fill=dbwrru,fill opacity=0.1] (2,0) -- (5,0) -- (5,-2) -- (2,-2) -- cycle;
\fill[line width=2pt,color=dbwrru,fill=dbwrru,fill opacity=0.1] (-4,-2) -- (-4,-4) -- (-1,-4) -- (-1,-2) -- cycle;
\fill[line width=2pt,color=dbwrru,fill=dbwrru,fill opacity=0.1] (-1,-4) -- (2,-4) -- (2,-2) -- (-1,-2) -- cycle;
\fill[line width=2pt,color=dbwrru,fill=dbwrru,fill opacity=0.1] (2,-2) -- (5,-2) -- (5,-4) -- (2,-4) -- cycle;
\draw [line width=2pt,color=dbwrru] (-4,2)-- (-4,0);
\draw [line width=2pt,color=dbwrru] (-4,0)-- (-1,0);
\draw [line width=2pt,color=dbwrru] (-1,0)-- (-1,2);
\draw [line width=2pt,color=dbwrru] (-1,2)-- (-4,2);
\draw [line width=2pt,color=dbwrru] (-1,2)-- (2,2);
\draw [line width=2pt,color=dbwrru] (2,2)-- (2,0);
\draw [line width=2pt,color=dbwrru] (2,0)-- (-1,0);
\draw [line width=2pt,color=dbwrru] (-1,0)-- (-1,2);
\draw [line width=2pt,color=dbwrru] (2,2)-- (5,2);
\draw [line width=2pt,color=dbwrru] (5,2)-- (5,0);
\draw [line width=2pt,color=dbwrru] (5,0)-- (2,0);
\draw [line width=2pt,color=dbwrru] (2,0)-- (2,2);
\draw [line width=2pt,color=dbwrru] (-4,0)-- (-4,-2);
\draw [line width=2pt,color=dbwrru] (-4,-2)-- (-1,-2);
\draw [line width=2pt,color=dbwrru] (-1,-2)-- (-1,0);
\draw [line width=2pt,color=dbwrru] (-1,0)-- (-4,0);
\draw [line width=2pt,color=dbwrru] (-1,-2)-- (2,-2);
\draw [line width=2pt,color=dbwrru] (2,-2)-- (2,0);
\draw [line width=2pt,color=dbwrru] (2,0)-- (-1,0);
\draw [line width=2pt,color=dbwrru] (-1,0)-- (-1,-2);
\draw [line width=2pt,color=dbwrru] (2,0)-- (5,0);
\draw [line width=2pt,color=dbwrru] (5,0)-- (5,-2);
\draw [line width=2pt,color=dbwrru] (5,-2)-- (2,-2);
\draw [line width=2pt,color=dbwrru] (2,-2)-- (2,0);
\draw [line width=2pt,color=dbwrru] (-4,-2)-- (-4,-4);
\draw [line width=2pt,color=dbwrru] (-4,-4)-- (-1,-4);
\draw [line width=2pt,color=dbwrru] (-1,-4)-- (-1,-2);
\draw [line width=2pt,color=dbwrru] (-1,-2)-- (-4,-2);
\draw [line width=2pt,color=dbwrru] (-1,-4)-- (2,-4);
\draw [line width=2pt,color=dbwrru] (2,-4)-- (2,-2);
\draw [line width=2pt,color=dbwrru] (2,-2)-- (-1,-2);
\draw [line width=2pt,color=dbwrru] (-1,-2)-- (-1,-4);
\draw [line width=2pt,color=dbwrru] (2,-2)-- (5,-2);
\draw [line width=2pt,color=dbwrru] (5,-2)-- (5,-4);
\draw [line width=2pt,color=dbwrru] (5,-4)-- (2,-4);
\draw [line width=2pt,color=dbwrru] (2,-4)-- (2,-2);
\draw [line width=2pt] (-2.48,1)-- (0.64,1);
\draw [line width=2pt] (0.64,1)-- (3.58,1);
\draw [line width=2pt] (3.58,1)-- (3.58,-1);
\draw [line width=2pt] (-2.48,1)-- (-2.48,-1);
\draw [line width=2pt] (-2.48,-1)-- (-2.46,-3);
\draw [line width=2pt] (-2.46,-3)-- (3.58,-3);
\draw [line width=2pt] (3.58,-1)-- (3.58,-3);
\draw [line width=2pt] (0.64,1)-- (0.6703308977726765,-3);
\draw [line width=2pt] (-2.48,-1)-- (3.58,-1);
\draw [line width=2pt,dash pattern=on 1pt off 1pt on 1pt off 4pt] (3.58,-1)-- (6,-1);
\draw [line width=2pt,dash pattern=on 1pt off 1pt on 1pt off 4pt] (3.58,1)-- (6,1);
\draw [line width=2pt,dash pattern=on 1pt off 1pt on 1pt off 4pt] (3.58,1)-- (3.6,3);
\draw [line width=2pt,dash pattern=on 1pt off 1pt on 1pt off 4pt] (0.64,1)-- (0.62,3);
\draw [line width=2pt,dash pattern=on 1pt off 1pt on 1pt off 4pt] (3.58,-3)-- (6,-3);
\draw [line width=2pt,dash pattern=on 1pt off 1pt on 1pt off 4pt] (3.58,-3)-- (3.58,-5.08);
\draw [line width=2pt,dash pattern=on 1pt off 1pt on 1pt off 4pt] (0.6703308977726765,-3)-- (0.68,-5.04);
\draw [line width=2pt,dash pattern=on 1pt off 1pt on 1pt off 4pt] (-2.46,-3)-- (-2.46,-5);
\draw [line width=2pt,dash pattern=on 1pt off 1pt on 1pt off 4pt] (-2.46,-3)-- (-5.16,-3);
\draw [line width=2pt,dash pattern=on 1pt off 1pt on 1pt off 4pt] (-2.48,-1)-- (-5,-1);
\draw [line width=2pt,dash pattern=on 1pt off 1pt on 1pt off 4pt] (-2.48,1)-- (-5.16,1);
\draw [line width=2pt,dash pattern=on 1pt off 1pt on 1pt off 4pt] (-2.48,1)-- (-2.48,3);
\begin{scriptsize}
\draw [fill=rvwvcq] (-2.48,1) circle (5pt);
\draw [fill=rvwvcq] (0.64,1) circle (5pt);
\draw [fill=rvwvcq] (3.58,1) circle (5pt);
\draw [fill=rvwvcq] (3.58,-1) circle (5pt);
\draw [fill=rvwvcq] (-2.48,-1) circle (5pt);
\draw [fill=rvwvcq] (0.64,-1) circle (5pt);
\draw [fill=rvwvcq] (-2.46,-3) circle (5pt);
\draw [fill=rvwvcq] (3.58,-3) circle (5pt);
\draw [fill=rvwvcq] (0.6703308977726765,-3) circle (5pt);
\end{scriptsize}
\end{tikzpicture}}
\caption{Geohash tessellation}
\end{subfigure}%
\begin{subfigure}{0.5\textwidth}
\centering
\scalebox{0.45}{
\begin{tikzpicture}[line cap=round,line join=round,>=triangle 45,x=1cm,y=1cm]
\fill[line width=2pt,color=dbwrru,fill=dbwrru,fill opacity=0.1] (0,3) -- (-1.47,2.24) -- (-1.71,0.22) -- (0.67,-0.5) -- (1.45,1.54) -- cycle;
\fill[line width=2pt,color=dbwrru,fill=dbwrru,fill opacity=0.1] (-1.71,0.22) -- (-3.47,0.74) -- (-2.45,-1.92) -- (0.21,-1.96) -- (0.67,-0.5) -- cycle;
\fill[line width=2pt,color=dbwrru,fill=dbwrru,fill opacity=0.1] (0.21,-1.96) -- (1.53,-3.54) -- (4.03,-3.2) -- (4.03,-1.28) -- (2.59,0) -- (0.67,-0.5) -- cycle;
\fill[line width=2pt,color=dbwrru,fill=dbwrru,fill opacity=0.1] (0.67,-0.5) -- (1.45,1.54) -- (4.05,2.16) -- (2.59,0) -- cycle;
\fill[line width=2pt,color=dbwrru,fill=dbwrru,fill opacity=0.1] (-2.45,-1.92) -- (-1.01,-3.62) -- (1.53,-3.54) -- (0.21,-1.96) -- cycle;
\draw [line width=2pt,color=dbwrru] (0,3)-- (-1.47,2.24);
\draw [line width=2pt,color=dbwrru] (-1.47,2.24)-- (-1.71,0.22);
\draw [line width=2pt,color=dbwrru] (-1.71,0.22)-- (0.67,-0.5);
\draw [line width=2pt,color=dbwrru] (0.67,-0.5)-- (1.45,1.54);
\draw [line width=2pt,color=dbwrru] (1.45,1.54)-- (0,3);
\draw [line width=2pt,color=dbwrru] (-1.71,0.22)-- (-3.47,0.74);
\draw [line width=2pt,color=dbwrru] (-3.47,0.74)-- (-2.45,-1.92);
\draw [line width=2pt,color=dbwrru] (-2.45,-1.92)-- (0.21,-1.96);
\draw [line width=2pt,color=dbwrru] (0.21,-1.96)-- (0.67,-0.5);
\draw [line width=2pt,color=dbwrru] (0.67,-0.5)-- (-1.71,0.22);
\draw [line width=2pt,color=dbwrru] (0.21,-1.96)-- (1.53,-3.54);
\draw [line width=2pt,color=dbwrru] (1.53,-3.54)-- (4.03,-3.2);
\draw [line width=2pt,color=dbwrru] (4.03,-3.2)-- (4.03,-1.28);
\draw [line width=2pt,color=dbwrru] (4.03,-1.28)-- (2.59,0);
\draw [line width=2pt,color=dbwrru] (2.59,0)-- (0.67,-0.5);
\draw [line width=2pt,color=dbwrru] (0.67,-0.5)-- (0.21,-1.96);
\draw [line width=2pt,color=dbwrru] (0.67,-0.5)-- (1.45,1.54);
\draw [line width=2pt,color=dbwrru] (1.45,1.54)-- (4.05,2.16);
\draw [line width=2pt,color=dbwrru] (4.05,2.16)-- (2.59,0);
\draw [line width=2pt,color=dbwrru] (2.59,0)-- (0.67,-0.5);
\draw [line width=2pt,color=dbwrru] (-2.45,-1.92)-- (-1.01,-3.62);
\draw [line width=2pt,color=dbwrru] (-1.01,-3.62)-- (1.53,-3.54);
\draw [line width=2pt,color=dbwrru] (1.53,-3.54)-- (0.21,-1.96);
\draw [line width=2pt,color=dbwrru] (0.21,-1.96)-- (-2.45,-1.92);
\draw [line width=2pt] (-0.11,1.38)-- (-1.65,-0.82);
\draw [line width=2pt] (-1.65,-0.82)-- (-0.35,-2.8);
\draw [line width=2pt] (-0.35,-2.8)-- (2.29,-1.78);
\draw [line width=2pt] (2.29,-1.78)-- (2.21,0.84);
\draw [line width=2pt] (2.21,0.84)-- (-0.11,1.38);
\draw [line width=2pt] (-1.65,-0.82)-- (2.29,-1.78);
\draw [line width=2pt,dash pattern=on 1pt off 1pt on 1pt off 4pt] (-0.35,-2.8)-- (0.35,-4.7);
\draw [line width=2pt,dash pattern=on 1pt off 1pt on 1pt off 4pt] (-0.35,-2.8)-- (-2.93,-3.18);
\draw [line width=2pt,dash pattern=on 1pt off 1pt on 1pt off 4pt] (-1.65,-0.82)-- (-4.61,-0.76);
\draw [line width=2pt,dash pattern=on 1pt off 1pt on 1pt off 4pt] (-1.65,-0.82)-- (-3.61,2.28);
\draw [line width=2pt,dash pattern=on 1pt off 1pt on 1pt off 4pt] (-0.11,1.38)-- (-0.75,4.34);
\draw [line width=2pt,dash pattern=on 1pt off 1pt on 1pt off 4pt] (-0.11,1.38)-- (-2.17,1.86);
\draw [line width=2pt,dash pattern=on 1pt off 1pt on 1pt off 4pt] (-0.11,1.38)-- (1.41,2.62);
\draw [line width=2pt,dash pattern=on 1pt off 1pt on 1pt off 4pt] (2.21,0.84)-- (2.21,2.64);
\draw [line width=2pt,dash pattern=on 1pt off 1pt on 1pt off 4pt] (2.21,0.84)-- (4.77,0.36);
\draw [line width=2pt,dash pattern=on 1pt off 1pt on 1pt off 4pt] (2.29,-1.78)-- (3.09,-4.22);
\draw [line width=2pt,dash pattern=on 1pt off 1pt on 1pt off 4pt] (2.29,-1.78)-- (4.93,-2.26);
\draw [line width=2pt,dash pattern=on 1pt off 1pt on 1pt off 4pt] (2.29,-1.78)-- (3.67,-0.6);
\begin{scriptsize}
\draw [fill=rvwvcq] (-0.11,1.38) circle (5pt);
\draw [fill=rvwvcq] (-1.65,-0.82) circle (5pt);
\draw [fill=rvwvcq] (-0.35,-2.8) circle (5pt);
\draw [fill=rvwvcq] (2.29,-1.78) circle (5pt);
\draw [fill=rvwvcq] (2.21,0.84) circle (5pt);
\end{scriptsize}
\end{tikzpicture}}
\caption{Voronoi tessellation}
\end{subfigure}
\vspace{0.3cm}
\caption{Representation of the Geohash and Voronoi tessellations as graphs. For each graph, the nodes are the centroids of the partitions. and the edges are the distances between the centroids. These graphical representations facilitate the application of the GraphLSTM on the spatial partitions. }
\label{geogebra}
\end{figure*}
\section{Spatio-Temporal Models} \label{modeling}
In the previous section, we saw that the spatial distribution of the data in each partition varies significantly with the partitioning technique employed (Fig.~\ref{heatmaps}). In this section, we examine whether this variation in spatial distribution has an impact on the performance of the prediction models employed in these partitions. Both the ConvLSTM and GraphLSTM models derive heavily from the LSTM network \cite{gers2000learning}. The RNN cell, from which the LSTM is developed, considers its present input and the output of the RNN cells preceding it, for its present output. The LSTM network overcomes several shortcomings of the plain RNN and learns long-term temporal dependencies. This property makes it a suitable candidate for time-series analysis. An LSTM cell has four NN units, called \emph{gates}, that interact with each other. See the equations below:
\begin{align}
& f_{t} = \delta(W_{f}\cdot X_{t}+R_{f}\cdot h_{t-1}+b_{f}), \label{1}\\
& i_{t} = \delta(W_{i} \cdot X_{t}+R_{i} \cdot h_{t-1}+b_{i}), \label{2}\\
& o_{t} = \delta(W_{o} \cdot X_{t}+R_{o} \cdot h_{t-1}+b_{o}), \label{3}\\
& \overline{C}_{t} = \varphi(W_{c} \cdot X_{t}+R_{c} \cdot h_{t-1}+b_{c}), \label{4}\\
& C_{t} = f_{t} \bullet C_{t-1}+i_{t}\bullet\overline{C}_{t}, \label{5}\\
& h_{t} = o_{t}\bullet\phi(C_{t}) \label{6},
\end{align}
where $X_t$ is the input and $W_{x}$, $R_{x}$, and $b_{x}$ represent the input weights, recurrent weights and bias of the gate $x$ respectively. The forget gate, given by Eqn.~(\ref{1}), decides the amount of historical information to be discarded. The input gate in Eqn.~(\ref{2}) decides the values to be updated in the cell state, and the output gate outputs the cell states in Eqn.~(\ref{3}).  The nonlinear activation functions $\delta$, $\varphi$ and $\phi$ squish the outputs to recommended ranges, which are usually [0,1] or [-1,1]. Matrix multiplication and element-wise product operations are denoted by $\cdot$ and $\bullet$ operators respectively. Eqn.~(\ref{4}) calculates a set of new candidate values to be added to the present cell state $C_{t}$. After the cell state is updated using Eqn.~(\ref{5}), the new hidden state output is given by Eqn.~(\ref{6}). 

\subsection{Geohash-based ConvLSTM} \label{clstm}
The Convolutional LSTM (ConvLSTM) network \cite{xingjian2015convolutional} combines the aspects of both CNN and LSTM. It extends a fully connected LSTM network to have convolutional structures in both input-to-state and state-to-state transitions, to learn spatial dependencies. The key equations are as follows: 
\begin{align}
& f_{t} = \delta(W_{f} * X_{t}+R_{f}* h_{t-1}+b_{f}), \\
& i_{t} = \delta(W_{i} * X_{t}+R_{i} * h_{t-1}+b_{i}), \\
& o_{t} = \delta(W_{o} * X_{t}+R_{o} * h_{t-1}+b_{o}),  \\
& \overline{C}_{t} = \varphi(W_{c} * X_{t}+R_{c} * h_{t-1}+b_{c}),\\
& C_{t} = f_{t} \bullet C_{t-1}+i_{t} \bullet \overline{C}_{t},\\
& h_{t} = o_{t} \bullet \phi(C_{t}),
\end{align}
where, the convolution operator is denoted by $*$. In mathematical terms, a ConvLSTM replaces the matrix multiplication operations in the feed-forward equations of the vanilla LSTM to convolution operations. If we consider the centroid of each partition as a node on a graph, the entire city can be represented by an undirected graph $G$. Each node represents a partition and will hold a value equal to the aggregated demand or supply for that partition. See Fig.~\ref{geogebra} for cross-sections of the graphs obtained using the Voronoi and Geohash schemes. The Voronoi tessellated city will generate an irregular graph, where all the nodes need not have the same number of neighbors. We note that the number of neighbors varies from 3 to 10 for each Voronoi partition. On the other hand, the Geohash tessellated city forms a highly regular graph where each node has 8 neighbors, equidistant from each other. This fixed structure of a Geohash-based graph allows us to apply standard convolution operations and hence, can be modeled using a standard ConvLSTM. For an arbitrarily structured graph like the Voronoi-based graph, localized rectangular filter operations cannot be applied.  A graph-based LSTM that utilizes the adjacency matrix to depict the structure of a graph can be employed in such scenarios. 
\subsection{Voronoi-based GraphLSTM}
The primary step of a GraphLSTM framework \cite{cui2018high} is to define the neighborhood. A k-hop neighborhood can be used to gather information from nodes that are k hops away from a node of interest. In this study, we gather spatial information from the first-order neighbors, \emph{i.e.,} the neighbors who share a common boundary with the partition of interest. Hence, a 1-hop neighborhood is considered for the implementation of our GraphLSTM. The 1-hop neighborhood matrix for any graph $\mathcal{G}$ is same as its adjacency matrix $\mathcal{A}$. To make the nodes self-accessible in the graph, the identity matrix $I$ is added to $\mathcal{A}$, to form $\Tilde{\mathcal{A}}$. Then, the 1-hop graph convolution at time $t$ can be defined as follows:
\begin{equation}
    \mathcal{GC}_t = (W_{gc} \bullet \Tilde{\mathcal{A}})X_t
\end{equation}
where, $W_{gc}$ is the 1-hop weight matrix for the 1-hop adjacency matrix, and $X_t \in \mathbb{R}^{N\times 1}$ is the demand or supply at time $t$, where $N$ is the number of Voronoi partitions. The features extracted from the graph convolution $\mathcal{GC}$ are fed to the LSTM network. We see that the structures of the forget gate $f_t$, the input gate $i_t$, the output gate $o_t$, and the input cell state $\overline{C}_{t}$ at time $t$ are similar to the vanilla LSTM.  The input is replaced by the graph convolution features $\mathcal{GC}$. A new cell state ${C^*}_{t-1}$ to incorporate the contributions of neighboring cell states is added to the framework, where $W_N$ is the corresponding weight matrix. The main equations are as follows:
\begin{align}
& f_{t} = \delta(W_{f}\cdot \mathcal{GC}_{t}+R_{f}\cdot h_{t-1}+b_{f}), \label{7}\\
& i_{t} = \delta(W_{i} \cdot \mathcal{GC}_{t}+R_{i} \cdot h_{t-1}+b_{i}), \label{8}\\
& o_{t} = \delta(W_{o} \cdot \mathcal{GC}_{t}+R_{o} \cdot h_{t-1}+b_{o}), \label{9} \\
& \overline{C}_{t} = \varphi(W_{c} \cdot \mathcal{GC}_{t}+R_{c} \cdot h_{t-1}+b_{c}), \label{10}\\
& {C^*}_{t-1} = W_N \bullet \Tilde{\mathcal{A}} \cdot C_{t-1}, \label{11}\\
& C_{t} = f_{t}\bullet{C^*}_{t-1}+i_{t}\bullet \overline{C}_{t},\label{12}\\
& h_{t} = o_{t}\bullet \text{tanh}(C_{t}).\label{13}
\end{align}
With the addition of ${C^*}_{t-1}$ in Eqn. (\ref{12}), the influence of the neighboring cell states will be considered during the recurrent updates of the cell state. We, then, compare the ConvLSTM and GraphLSTM networks against LSTM networks modeled using Voronoi and Geohash features. The ARIMA and ARIMAX models are also used as baselines to explore the assumption that the relationship between the data in adjacent partitions is linear.

\subsection{LSTM}
For a region $r_i$ (Voronoi cell or geohash), we first feed the demand/supply from $r_i$ alone to the LSTM network. Then, to analyze the effect of spatial neighbors, we feed data from $r_i$ and its first-order neighbors to the LSTM network. We vary the number of first-order neighbors to arrive at the best spatial configuration.

\subsection{ARIMA and ARIMAX}
For linear modeling, we consider two models: (i) ARIMA (Auto Regressive Integrated Moving Average), and (ii) ARIMAX (ARIMA with  eXogenous inputs). After examining several regression models, we observed that ARIMA is a satisfactory fit for a majority of Voronoi cells and geohashes. Hence, we aim to fit a single ARIMA and ARIMAX model for the entire city. The ARIMA model belongs to the class of statistical modeling, with an Auto Regressive (AR) part to model the changing variable as a regression on its own lagged values, an Integrated (I) part to produce stationary series, and a Moving Average (MA) part to incorporate the dependency between an observation and the residual errors obtained from a moving average model applied to lagged observations. The model is represented as:
\begin{equation}
  y'_{t} =  \phi_{1}y'_{t-1} + \cdots + \phi_{p}y'_{t-p}
     + \theta_{1}\varepsilon_{t-1} + \cdots + \theta_{q}\varepsilon_{t-q} + \varepsilon_{t},
\end{equation}
where $y_t$ is the demand/supply to be predicted at time t, $y'_{t}$ is the differenced form of $y_t$, $p$ and $\phi$ are the order and parameters of the AR process, $q$ and $\theta$ are the order and parameters of the MA process, and $\varepsilon_t$ is the forecast error. Historical information from the variable of interest alone is taken into consideration for the standard ARIMA model. The ARIMAX model is an extension of ARIMA that provides a framework to include information from the neighboring regions (\emph{i.e.,} covariates). We employ the ARIMAX to analyze the extent of spatial information captured with different tessellation schemes. The ARIMAX model is defined as:
\begin{equation}
    y'_t = \beta z_t + \phi_{1}y'_{t-1} + \cdots + \phi_{p}y'_{t-p}
     + \theta_{1}\varepsilon_{t-1} + \cdots + \theta_{q}\varepsilon_{t-q} + \varepsilon_{t},
\end{equation}
where $z_t$ is the covariate at time $t$, and the parameter $\beta$ includes the lagged versions of the covariate. The time-sequences from the positively correlated first-order neighboring regions serve as covariates in our study.
\begin{table}[t!]
\centering
{\renewcommand{\arraystretch}{1.1}
\begin{tabular}{|l|}
\hline
\multicolumn{1}{|c|}{{Range of Hyper-parameters that are fed to TPE-BO}}                                                                                                                          \\ \hline \hline
Number of layers, L = [1, 2]                                                                                                                                 \\ \hline
Number of neurons, n = [10, 20, 50, 100]                                                                                                                     \\ \hline
Dropout, D = Uniform (0,0.5)     \\ \hline
Activation = [Sigmoid, Relu, Linear]                                                                                                                         \\ \hline
Optimizer = [Adam, Stochastic Gradient, RMSprop]                                                                                                             \\ \hline
Learning Rate = [\scalebox{0.9}{$10^{-1}, 10^{-2}, 10^{-3}, 10^{-4}, 10^{-5}, 10^{-6}$}]\\ \hline
Batch Size = [64, 128]                                                                                                                                        \\ \hline
\end{tabular}}
\caption{The optimal values obtained from this range of hyper-parameters, after performing the TPE (Tree-structured Parzen Estimator) based BO (Bayesian Optimization), are used for modeling the taxi demand-supply data.}
\label{parameters}
\end{table}
\begin{table*}[htb!]
\centering
\scalebox{0.72}{
{\renewcommand{\arraystretch}{1.6}
\begin{tabular}{|c|c|c|c|c||c|c|c||c|c|c|}
\hline
\multicolumn{2}{|c|}{\multirow{2}{*}{\begin{tabular}[c|]{@{}c@{}}Spatio-Temporal \\ Models\end{tabular}}} & \multicolumn{3}{c||}{Bengaluru Demand} & \multicolumn{3}{c||}{Bengaluru Supply} & \multicolumn{3}{c|}{NYC Demand} \\ \cline{3-11} 
\multicolumn{2}{|c|}{}  & MASE       & SMAPE       & RMSE       & MASE       & SMAPE       & RMSE       & MASE     & SMAPE     & RMSE     \\ \hline \hline
\multirow{2}{*}{ARIMA}  & {G} &  1.31 & 48.5 & 12.36  &   18.0 & 191.1 & 57.3 &  1213.6  &  169.8 & 447.2 \\[1ex] \cline{2-11} 
                        & {V} &    1.13 & 43.5 & 8.14 & 4.68 &  95.4 & 15.8 & 5.01 & 125.5 & 27.6  \\[1ex]  \hline

\multirow{2}{*}{ARIMAX}  & G & \begin{tabular}[c]{@{}c@{}}1.13\\  (1)\end{tabular}  & \begin{tabular}[c]{@{}c@{}}43.4\\  (1)\end{tabular} &   \begin{tabular}[c]{@{}c@{}}10.1\\  (1)\end{tabular}       & \begin{tabular}[c]{@{}c@{}}1.34\\ (8)\end{tabular}    & \begin{tabular}[c]{@{}c@{}}60.1 \\ (8)\end{tabular}  & \begin{tabular}[c]{@{}c@{}}9.66\\ (8)\end{tabular}             & \begin{tabular}[c]{@{}c@{}}2.42 \\ (4)\end{tabular}  & \begin{tabular}[c]{@{}c@{}}80.1\\  (2)\end{tabular}   & \begin{tabular}[c]{@{}c@{}}123.3 \\ (2)\end{tabular}  \\[-0.75ex]  \cline{2-11}
        & V     & \begin{tabular}[c]{@{}c@{}}1.20\\ (8)\end{tabular}    & \begin{tabular}[c]{@{}c@{}}47.2 \\ (8)\end{tabular}   & \begin{tabular}[c]{@{}c@{}}8.93\\  (8)\end{tabular}       
        & \begin{tabular}[c]{@{}c@{}}1.31\\  (2)\end{tabular}            & \begin{tabular}[c]{@{}c@{}}58.7 \\ (7)\end{tabular}  & \begin{tabular}[c]{@{}c@{}}8.03 \\ (7)\end{tabular}       
        & \begin{tabular}[c]{@{}c@{}}1.84 \\ (4)\end{tabular}           & \begin{tabular}[c]{@{}c@{}}94.0 \\ (4)\end{tabular} & \begin{tabular}[c]{@{}c@{}}15.6 \\ (4)\end{tabular}           \\[-0.75ex] \hline

\multirow{3}{*}{LSTM}   & G & \begin{tabular}[c]{@{}c@{}}0.75 $\pm$ 0.26 \\ (7)\end{tabular} & \begin{tabular}[c]{@{}c@{}}16.14 $\pm$ 4.99\\ (7)\end{tabular}  & 
\begin{tabular}[c]{@{}c@{}}6.45 $\pm$ 4.51\\ (7)\end{tabular}  & \begin{tabular}[c]{@{}c@{}}0.93 $\pm$ 0.34\\ (6)\end{tabular}  & \begin{tabular}[c]{@{}c@{}}21.5 $\pm$ 5.65\\ (4)\end{tabular} & \begin{tabular}[c]{@{}c@{}}7.11 $\pm$ 6.17\\ (4)\end{tabular} & \begin{tabular}[c]{@{}c@{}}0.85 $\pm$ 0.41\\ (8)\end{tabular} & \begin{tabular}[c]{@{}c@{}}23.2 $\pm$ 40.9\\ (4)\end{tabular} & \begin{tabular}[c]{@{}c@{}}58.9 $\pm$ 34.5\\ (8)\end{tabular} \\[-0.75ex]   \cline{2-11} 
        & V     & \begin{tabular}[c]{@{}c@{}}0.76 $\pm$ 0.16 \\ (6)\end{tabular}    & \begin{tabular}[c]{@{}c@{}}16.72 $\pm$ 3.63 \\ (6)\end{tabular}   & \begin{tabular}[c]{@{}c@{}}5.45 $\pm$ 2.42\\ (6)\end{tabular}     & \begin{tabular}[c]{@{}c@{}}0.98 $\pm$ 0.29\\  (2)\end{tabular}    & \begin{tabular}[c]{@{}c@{}}28.1 $\pm$ 30.8\\(7)\end{tabular} & \begin{tabular}[c]{@{}c@{}}6.68 $\pm$ 5.37\\  (2)\end{tabular}  & \begin{tabular}[c]{@{}c@{}}0.88 $\pm$ 0.19\\ (8)\end{tabular}     & \begin{tabular}[c]{@{}c@{}}29.3 $\pm$ 104\\ (0)\end{tabular}     & \begin{tabular}[c]{@{}c@{}}8.26 $\pm$ 5.73\\ (8)\end{tabular} \\[-0.75ex]  \hline
     
ConvLSTM    & G     & 0.37 $\pm$ 1.99                   & \textbf{9.1 $\pm$ 4.8}    & \textbf{2.16 $\pm$ 1.52}      & 1.51 $\pm$ 2.23                                  & 35.2 $\pm$ 12.7                   & \textbf{7.73 $\pm$  2.03} & 40.5 $\pm$ 0.05               & 12.8 $\pm$ 15.4                                  & 36.8 $\pm$ 10.4                                               \\ [1ex]\hline 
\multirow{2}{*}{GraphLSTM}      & V                 & \textbf{0.72 $\pm$ 0.15}  & 16.1 $\pm$ 3.66               & 4.99 $\pm$ 2.35                                                  & 0.93 $\pm$ 0.50   & \textbf{21.8 $\pm$ 4.77}  & 6.15 $\pm$ 4.96               & \textbf{0.68 $\pm$ 0.16}                                         & 17.4 $\pm$ 4.32   & \textbf{6.33 $\pm$ 4.51}                                               \\[1ex] \cline{2-11} 
                                    & G                 & 0.73 $\pm$ 0.20           & 15.6 $\pm$ 5.1                & 6.2 $\pm$ 4.75            
                                    &\textbf{0.92 $\pm$ 0.38}  & 20.7 $\pm$ 5.9     & 6.79 $\pm$ 5.91               & 0.71 $\pm$ 0.68           
                                    & \textbf{11.9 $\pm$ 4.56} & 48.2 $\pm$ 29.3                                               \\[1ex] \hline \hline
\multicolumn{2}{|c|}{\scalebox{1.2}Proposed}        & \scalebox{1.2}{\textbf{0.32 $\pm$ 0.90}}   & \scalebox{1.2}{\textbf{8.5 $\pm$ 3.7}}             & \scalebox{1.2}{\textbf{2.25 $\pm$ 1.90}}               & \scalebox{1.2}{\textbf{0.81 $\pm$ 0.96}}                                                  & \scalebox{1.2}{\textbf{17.7 $\pm$ 4.51}}   & \scalebox{1.2}{\textbf{4.61 $\pm$ 2.73}}           & \scalebox{1.2}{\textbf{0.43 $\pm$ 0.17}}               & \scalebox{1.2}{\textbf{8.90 $\pm$ 3.6}}                                                   & \scalebox{1.2}{\textbf{6.48 $\pm$ 5.5}}                                \\[1ex] \hline
\end{tabular}}}
\caption{Predictive performance of various models across data sets with different performance metrics. The numbers in braces associated with ARIMAX and LSTM models denote the number of spatial features that resulted in the best performance. For the NN models, the errors are given as \emph{mean $\pm$ standard deviation}. The GraphLSTM shows an overall robust performance, and the proposed dHEDGE based prediction technique ensures consistently good performance across different scenarios.}
\label{resultstable}
\end{table*}
\section{Experiments}\label{experiments}
In this section, we discuss the hyper-parameter optimization techniques for the models, evaluation metrics, and inferences obtained on performing the comparison study. 
\subsection{Experimental settings}
Before modeling the data using any NN model, it is necessary to set the optimal hyper-parameters. Hyper-parameters are the model-specific properties that are to be fixed before the training phase of the model. They define the high-level properties of the model such as the time complexity or the learning rate. Out of the various algorithms available for hyper-parameter optimization, Bayesian Optimization is widely used in the recent machine learning literature \cite{shahriari2016taking}. In this study, we use the Tree-structured Parzen Estimator Bayesian Optimization (TPE-BO) \cite{bergstra2011algorithms} approach for tuning the hyper-parameters. This algorithm uses Parzen estimators to model the error distribution as non-parametric densities. The range of the hyper-parameters fed to the TPE algorithm is given in Table \ref{parameters}. Additionally, for ConvLSTM, we vary the number of filters from 16 to 258. The RMSprop is shortlisted as the optimization function for our data sets. The choice of the activation function at the output dense layer is Relu. The Relu activation is recommended for data sets such as passenger count and taxi supply as it allows the output to vary linearly, with a minimum at zero. That is, the output is zero if the input to the activation function is less than zero. In case the input is greater than zero, the output is equal to the input. The dropout values, number of layers and learning rates are optimized for each data set, tessellation technique, and NN model. In addition to the hyper-parameter values suggested by the TPE-BO, we manually tune the parameters to arrive at the best prediction accuracy.

The K-Means clustering algorithm generates a set of \emph{N} regions of interest. The variable \emph{N} takes values 740 and 780 for Bengaluru and New York City respectively. For each \emph{n} $\in$ \emph{N}, two time-sequences are generated using the Voronoi and Geohash tessellation strategies.  The data is aggregated over 60 minutes for 60 days, generating time-sequences of length \emph{T} = 1440 time steps. To implement GraphLSTM for Voronoi tessellation, we pick the 1-hop neighbors for every node \emph{n}. The GraphLSTM receives inputs of the form \textbf{X} $\in \mathbb{R}^{N \times T}$, along with an adjacency matrix $\Tilde{A} \in  \mathbb{R}^{N \times N}$.  The adjacency matrix encapsulates information from the first-order neighbors. For the GraphLSTM, we consider a hidden layer with dimension equal to the number of nodes in the graph. For ConvLSTM, we consider frames of size 3$\times$3. This particular configuration allows us to capture information from a 6-level geohash of interest $n$ (the center pixel) and 8 first-order neighbors.  The ConvLSTM framework receives inputs of the form \textbf{X} $\in \mathbb{R}^{N \times T \times 3 \times 3 \times 1}$, where frame sizes are of dimension 3 $\times$ 3, along a single channel. For the LSTM network, the inputs are of the form \textbf{X} $\in \mathbb{R}^{N \times T \times S}$, where \emph{S} is the number of spatial neighbors.  Note that while a geohash has 8 fixed number of first-order neighbors, a Voronoi cell has a variable number of first-order neighbors. This corresponds to a \emph{S} value of 8 for Geohash LSTM. For consistent comparison, while training Voronoi input based LSTM, we pick features from the top 8 positively correlated Voronoi neighbors.  For Voronoi cells with less than 8 neighbors, we compensate for the lack of features by introducing invalid feature vectors to differentiate them from useful information. 

For ARIMA and ARIMAX models, we vary the AR and MA parameters between the range [0, 5] and the time-sequences are differenced whenever non-stationary behavior is encountered.  While fitting the LSTM and ARIMAX models to the city, we varied the number of spatial Voronoi and Geohash features included in the model, to find the best spatial configurations for each data set. All the NN models are trained with a batch size of 64 for 500 epochs and repeated 5 times to compensate for the random initialization of network weights. MinMax scaling is applied to the input before they are fed to the various networks. Early stopping mechanism is employed to prevent over-fitting.

\subsection{Evaluation metrics}
For each data set, data from the first 59 days is used for training the models. The models are then tested on the $\text{60}^{th}$ day. We keep aside 10\% of the training data for validation purposes. We employ three widely used performance metrics to evaluate the models:
\begin{enumerate}
    \item Symmetric Mean Absolute Percentage Error (SMAPE):
    \begin{equation}
           \text{SMAPE} = \frac{100}{h} \mathlarger{\sum_{t=1}^{h}}\frac{|{y}_{t}-\hat{y}_{t}|}{\hat{y}_{t}+y_{t}+1},
    \end{equation}
    \item Mean Absolute Scaled Error (MASE):
    \begin{equation}
            \text{MASE} = \frac{1}{h} \mathlarger{\sum_{t= 1}^h}\Bigg (\frac{|y_t - \hat{y_t}|}{\frac{1}{n-m}\sum_{t=m+1}^n |y_t - y_{t-m}|} \Bigg ),
    \end{equation}
    \item Root Mean Square Error (RMSE):
    \begin{equation}
            \text{RMSE} = \sqrt{\frac{1}{h}{\sum_{t=1}^h (y_t - \hat{y_t})^2}},
    \end{equation}
\end{enumerate}
where $h$ is the forecast horizon, $m$ is the seasonal period, $y_t$ is the actual demand, $n$ is the length of the training set, and $\hat{y_t}$ is the forecast at time $t$. The RMSE gives relatively high weights to large errors. The SMAPE is an accuracy measure based on percentage errors. Both RMSE and SMAPE are scale dependent errors. The MASE compares the forecast errors of the test set with the in-sample forecast errors from the standard Na\"{i}ve model, making it scale independent. Since these performance metrics are based on the $L_1$-norm and $L_2$-norm errors, we define the loss function over which the optimization is performed as:  
\begin{equation}
    \begin{aligned}
      \text{Loss function, L} = & \ \frac{1}{N}\Bigg({\sum_{t=1}^N (y_t - \hat{y_t})^2} + {|y_t - \hat{y_t}|}\Bigg). \label{loss}
    \end{aligned}
\end{equation}
Since the loss function is the sum of the Mean Squared Error (MSE) and Mean Absolute Error (MAE), the models will optimize for both $L_1$ and $L_2$ errors. 

\subsection{Experimental results} \label{experimentalresults}
The Table \ref{resultstable} summarizes the numerical results of the comparison study. The results are obtained by applying the regression and NN models on the three data sets aggregated using Voronoi and Geohash tessellation schemes. The standard deviation factor accounts for the variability across different locations and multiple runs. The main inferences drawn from the study are as follows:
\begin{enumerate}
    \item Even though the overall performance of linear regression models is sub-par with that of the non-linear neural models, the high computational speed and comparable performance in certain scenarios are to be noted.
    \item The entire set of first-order neighbors may not be necessary to achieve the best spatial model configuration.
    \item The prediction performance of the GraphLSTM is better than that of the standard ConvLSTM on the majority of test cases, and this was achieved with lower computational complexity.
    \item Across data sets and metrics, the irregular Voronoi graph based GraphLSTM performs comparable to or better than the regular Geohash graph based GraphLSTM, suggesting at the potential of irregular graphs in location-based forecasting.  
    \item The lack of a universal winning tessellation strategy is noted. 
\end{enumerate}

The ARIMA model achieves better prediction accuracy with Voronoi tessellation based input features than with Geohash features. The overall accuracy improved on incorporating spatial information through ARIMAX models. However, we notice that ARIMAX models resulted in performance deterioration for Bengaluru demand data set. To investigate this behavior further, we modeled the top 50 demand scarce and demand dense regions in Bengaluru using independent ARIMAX models and saw clear improvements in accuracy. This points towards the inability of a linear regression model to satisfactorily capture spatial information using a single model for the entire city. Hence, with regression models, we do not recommend modeling the entire city with a single model. Voronoi and Geohash based LSTM models show consistent improvements in accuracy on incorporating information from spatial neighbors. We notice that all the first-order neighbors might not be required to arrive at the best spatio-temporal model configuration. 

The ConvLSTM model based on the Geohash strategy achieves good prediction accuracy on the Bengaluru data set but fails to perform well for the New York city data set. This trend is seen across deeper neural layers and different filter depths. On the other hand, Voronoi-based GraphLSTM exhibits consistently high prediction performance across multiple scenarios for both cities. The poor performance of Geohash-based ConvLSTM with NYC demand data can be attributed to the highly skewed spatial data distribution. 90\% of the total data is concentrated around the Manhattan borough, leaving the other four boroughs with 10\% of the total demand. Employing a fixed-sized partitioning scheme in a non-uniform data distributed space is not an efficient model setting. Geohash partitioning results in a large number of demand scarce cells in some boroughs, affecting model performance. Meanwhile, K-Means based Voronoi tessellation attempts to uniformly distribute data in the partitions, resulting in a lower number of demand scarce cells. GraphLSTM based on such an efficient model setting achieves high prediction performance. 

For further validation of this observation, we represent the Geohash partitions as nodes on a regular graph and conduct Geohash-based GraphLSTM modeling. We find that the Geohash-based GraphLSTM is also unable to model the data satisfactorily, resulting in high variability in the RMSE and MASE.  This highlights the importance of choosing an appropriate tessellation strategy, irrespective of the modeling technique used. In this case, the right choice of the partitioning technique resulted in  80\% improvement in RMSE. The Bengaluru demand and supply data sets have a uniform spatial distribution, and hence, a Geohash partitioning scheme based model works sufficiently well. Therefore, we conclude that the choice of the spatial partitioning technique depends on the data distribution. Even then, the Voronoi partitioning scheme based model exhibits competitive prediction performance at a lower computational cost. This shows that an appropriate tessellation strategy can reduce the complexity of the network to be built, without compromising on the prediction accuracy. The GraphLSTM has roughly the computational complexity of the vanilla LSTM, which is much lesser than that of the ConvLSTM. While the ConvLSTM has additional convolutional layers that increase the number of matrix operations, GraphLSTM builds on the vanilla LSTM architecture, with one additional gate and some changes to the input. Note that we measure the computational complexity in terms of the matrix operations to be performed. The lower error variance of the GraphLSTM in comparison with that of the ConvLSTM suggests that the consistency in predictions is also maintained across various locations in the city. To summarize, the consistent performance of GraphLSTM, combined with low computational complexity, across scenarios in the context of location-based passenger demand and driver supply forecasting merits attention and needs further exploration.  

\begin{figure}[t!]
\centering
        \psfrag{t}{\hspace{-10mm} \raisebox{-1.75mm}{\footnotesize{Forecast horizon}}}
        \psfrag{m}{\hspace{-5.5mm}\raisebox{1mm}{\footnotesize{Magnitude}}}
        \psfrag{s}{\hspace{0mm}\raisebox{-0.9mm}{\scalebox{0.7}{True Supply}}}
        \psfrag{d}{\hspace{0mm}\raisebox{-0.9mm}{\scalebox{0.7}{True Demand}}}
        \psfrag{sp}{\hspace{0mm}\raisebox{-0.9mm}{\scalebox{0.7}{Pred. Supply}}}
        \psfrag{dp}{\hspace{0mm}\raisebox{-0.9mm}{\scalebox{0.7}{Pred. Demand}}}
            	\psfrag{1}{\hspace{0mm}\raisebox{-0.5mm}{\footnotesize{1}}}
            	 \psfrag{12}{\hspace{0mm}\raisebox{-0.5mm}{\footnotesize{12}}}
            	\psfrag{24}{\hspace{0mm}\raisebox{-0.5mm}{\footnotesize{24}}}
            	\psfrag{0}{\hspace{0mm}\raisebox{-0.1mm}{\footnotesize{0}}}
            	\psfrag{30}{\hspace{-1mm}\raisebox{-0.1mm}{\footnotesize{30}}}
            	\psfrag{60}{\hspace{-2mm}\raisebox{-0.1mm}{\footnotesize{60}}}
        \includegraphics[height = 1.5in, width = 2.5in]{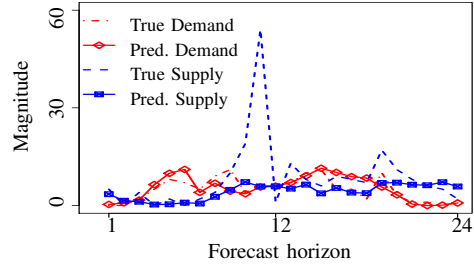}
    \caption{The actual and forecast demand-supply patterns for a demand scarce Voronoi cell. The supply forecast appears to a better representation of the supply required in that region than the actual supply levels.}
    \label{sdmatch}
 \end{figure}

While we stress on the significance of selecting an appropriate tessellation strategy while modeling, we observe that the best tessellation strategy and hence, the prediction model, varies with the choice of data set and performance metric. While the GraphLSTM has an overall favorable performance across data sets, there are instances where ConvLSTM outperforms GraphLSTM (e.g., in the Bengaluru demand data). For ensuring good prediction performance across all scenarios, we explore ensemble learning in the next section. Fig. \ref{sdmatch} shows the GraphLSTM based demand-supply predictions plotted against the actual patterns in a top Voronoi cell in Bengaluru. We see that the supply predicted from the historical data is a better match than the actual supply for the existing demand patterns in that region. Hence, a rerouting decision based on the predicted supply may reduce the demand-supply mismatch. 


\begin{figure}
    \centering
       \psfrag{gv}{\hspace{-1mm} \raisebox{0mm}{\scalebox{0.7}{V. GraphLSTM}}}
       \psfrag{gg}{\hspace{-1mm} \raisebox{0mm}{\scalebox{0.7}{G. GraphLSTM}}}
        \psfrag{cg}{\hspace{-1mm} \raisebox{0mm}{\scalebox{0.7}{G. ConvLSTM}}}
        \psfrag{mase}{\hspace{-4mm} \raisebox{0mm}{\scalebox{0.8}{MASE}}}
       \psfrag{h}{\hspace{-1mm} \raisebox{0mm}{\scalebox{0.7}{dHEDGE}}}
        \psfrag{t}{\hspace{-6mm}\raisebox{-1.75mm}{\footnotesize{Forecast Horizon}}}
        \psfrag{0.1}{\hspace{-2mm} \raisebox{0mm}{\footnotesize{0}}}
        \psfrag{2.6}{\hspace{-2.5mm} \raisebox{0mm}{\footnotesize{2.5}}}
       \psfrag{5.0}{\hspace{-1mm} \raisebox{0mm}{\footnotesize{5}}}
    \psfrag{1}{\hspace{-1mm} \raisebox{-0.5mm}{\footnotesize{1}}}    
    \psfrag{12}{\hspace{-1mm} \raisebox{-0.5mm}{\footnotesize{12}}}    
    \psfrag{24}{\hspace{-1mm} \raisebox{-0.5mm}{\footnotesize{24}}}
   \includegraphics[height = 1.5in,width = 2.5in]{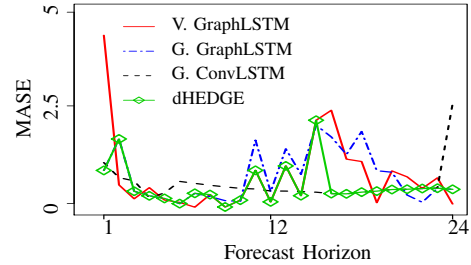}
    \caption{Performance of the online combining algorithm on the prediction models, where V. and G. denote Voronoi and Geohash respectively. The dHEDGE algorithm closely follows the best strategy at each time step in the forecast horizon.}
    \label{dhedgefig}
\end{figure}

\section{Combining models with dHEDGE algorithm}\label{dhedge}

The applicability of ensemble learning in a non-stationary environment that involves tessellation strategies was put forward in \cite{davis2018ataxi}, where we applied dHEDGE ensemble learning algorithm to combine the tessellation strategies. We observed that the regression models based on any one of the tessellation strategies were unable to yield optimal results for the entire forecast horizon. On exploring various LSTM networks, we note that this observation extends to the performance of RNN models as well, thereby strengthening the claim put forward in \cite{davis2018ataxi}. In Fig. \ref{dhedgefig}, we see that the best prediction technique varies with the time of the day. The best strategy switches between the Voronoi and Geohash tessellation based techniques throughout the forecast horizon. We find that irrespective of the modeling tool used, there is no universally superior tessellation strategy. This is in addition to the dependency of the strategies on the data set and performance metric (Table \ref{resultstable}). 

To compensate for the lack of a winner strategy, we use a variant of the well-known HEDGE algorithm \cite{freund1995desicion} suited for the non-stationary environment known as the dHEDGE algorithm \cite{raj2017aggregating}. By exponentially reducing the weights associated with each expert (\emph{i.e.,} prediction model), the dHEDGE takes into account the non-stationarity of the process. In our case, we have three experts, Voronoi and Geohash based GraphLSTM models, and the ConvLSTM model. The weight initialization can be performed either uniformly or based on some prior knowledge about the experts. We initialize the weights based on a holdout validation set. The weights are updated based on the previous weights, a discounting factor $\gamma$, a learning factor $\beta$, and a loss function $l_{t}$. The loss function $l_{t}$ is based on the prediction errors observed by the experts at time $t$. Thus, for each time step $t$, the weights are updated for the $i^{th}$ expert as:
\begin{equation}
    w_i[t+1] = w_i[t]^{\gamma} \cdot \beta^{l_i[t]}.
\end{equation}
The discounting and learning factors are chosen based on the validation set. The performance of the algorithm can be seen in Fig. \ref{dhedgefig}.  At each time step in the forecasting horizon, we pick the expert with the highest weight and use its predictions. We find that the algorithm picks the best shifting expert, by giving more weightage to the behavior of that expert in the recent past. The interested reader can refer to \cite{davis2018ataxi} for a detailed analysis of the algorithm. The prediction accuracy after applying the dHEDGE algorithm on the three experts can be seen in Table \ref{resultstable}. The algorithm consistently results in an accuracy close to the best expert in the pool. To demonstrate the flexibility of our algorithm in adapting to various scenarios, we evaluate the performance of our algorithm on the New York demand data set. While Voronoi GraphLSTM achieves good prediction accuracy, the two Geohash-based models perform poorly. When the dHEDGE is applied on these three experts, it is remarkable to note that dHEDGE achieves an accuracy close to the Voronoi GraphLSTM, and is unaffected by the poor performance of the other two experts. Further, in some use cases, we note that combining the experts produces accuracy levels better than that of any of the individual experts. This behavior is attributed to the time-dependent behavior of the models. In conclusion, our algorithm provides consistent performance across data sets and performance metrics, eliminating various dependencies of the prediction models.

\section{\ \ Concluding Remarks}\label{conclusions}
In the context of e-hailing taxi services, generating accurate demand-supply forecasts is instrumental in minimizing customer wait times and maximizing driver utilization. Neural Network-based taxi demand or supply forecasting commonly uses a fixed-sized equally-spaced spatial partitioning scheme. In this paper, we explored the impact of different spatial partitioning schemes on the predictive performance of LSTM (Long Short-Term Memory) models. By comparing ConvLSTM (Convolutional LSTM) and GraphLSTM (Graph-based LSTM), we draw attention to the potential of learning the partitioned city structure as a graph and applying Graph-based Neural Networks. 

When evaluated on three large-scale real-world data sets, GraphLSTM emerged as a promising candidate for location-based taxi demand-supply forecasting. The GraphLSTM offered competitive prediction performance against ConvLSTM at a much lower computational complexity. The comparison between GraphLSTM models based on regular and irregular graphs revealed the potential of irregular graphs in the context of location-based forecasting.  

In addition to the proposal to use irregular graph based GraphLSTM for taxi demand-supply forecasting, the findings in this paper recommend exploration and selection of a suitable tessellation strategy prior to fitting a Neural Network model. The choice of a suitable prediction model was found to depend on the properties of the data set and the performance metric employed. 

To achieve superior performance across all scenarios, we recommend the dHEDGE based ensemble learning algorithm. By employing dHEDGE in conjunction with the GraphLSTM and ConvLSTM models, we consistently achieved a prediction accuracy close to the best model at each time instant across the data sets considered, with different performance metrics.


\subsection{Avenues for further research}
This paper was directed towards accurate forecasting of taxi demand and supply, where we highlighted the potential of Graph-based LSTM networks. A detailed analysis of GraphLSTM can be performed, using more real-world data sets. Further, we note that demand-supply mismatches also occur when there are unexpected spikes in demand. In our future work, we aim to detect and include such anomalous events in the prediction model to achieve better predictions.




\begin{thebibliography}{10}

\bibitem{colborn2018spatio}
K.~L. Colborn, E.~Giorgi, A.~J. Monaghan, E.~Gudo, B.~Candrinho, T.~J. Marrufo,
  and J.~M. Colborn, ``Spatio-temporal modelling of weekly malaria incidence in
  children under 5 for early epidemic detection in mozambique,''
  \emph{Scientific Reports}, vol.~8, p. 9238, 2018.
  
\bibitem{ezzat2018spatio}
A.~A. Ezzat, M.~Jun, and Y.~Ding, ``Spatio-temporal asymmetry of local wind
  fields and its impact on short-term wind forecasting,'' \emph{IEEE
  Transactions on Sustainable Energy}, vol.~9, pp.~1437-1447, 2018.

\bibitem{wang2018spatio}
X.~Wang, Z.~Zhou, F.~Xiao, K.~Xing, Z.~Yang, Y.~Liu, and C.~Peng,
  ``Spatio-temporal analysis and prediction of cellular traffic in
  metropolis,'' \emph{IEEE Transactions on Mobile Computing (Early Access)}, 2018.
  
\bibitem{davis2018ataxi}
N.~Davis, G.~Raina, and K.~Jagannathan, ``Taxi demand forecasting: A
  HEDGE-based tessellation strategy for improved accuracy,'' \emph{IEEE
  Transactions on Intelligent Transportation Systems}, vol.~19, pp.
  3686--3697, 2018.

\bibitem{kamga2015analysis}
C.~Kamga, M.~A. Yazici, and A.~Singhal, ``Analysis of taxi demand and supply in
  New York City: implications of recent taxi regulations,''
  \emph{Transportation Planning and Technology}, vol.~38, pp. 601--625,
  2015.

\bibitem{jager2016analyzing}
B.~J{\"a}ger, M.~Wittmann, and M.~Lienkamp, ``Analyzing and modeling a city’s
  spatiotemporal taxi supply and demand: A case study for {M}unich,''
  \emph{Journal of Traffic and Logistics Engineering}, vol.~4, pp.~147--153, 2016.

\bibitem{tong2017simpler}
Y.~Tong, Y.~Chen, Z.~Zhou, L.~Chen, J.~Wang, Q.~Yang, J.~Ye, and W.~Lv, ``The
  simpler the better: a unified approach to predicting original taxi demands
  based on large-scale online platforms,'' in \emph{Proc. of the ACM
  SIGKDD International Conference on Knowledge Discovery and Data
  Mining}, pp.~1653--1662, 2017.

\bibitem{ke2017short}
J.~Ke, H.~Zheng, H.~Yang, and X.~M. Chen, ``Short-term forecasting of passenger
  demand under on-demand ride services: A spatio-temporal deep learning
  approach,'' \emph{Transportation Research Part C: Emerging Technologies},
  vol.~85, pp. 591--608, 2017.

\bibitem{nagy2018survey}
A.~M. Nagy and V.~Simon, ``Survey on traffic prediction in smart cities,''
  \emph{Pervasive and Mobile Computing}, vol.~50, pp. 148--163, 2018.

\bibitem{nguyen2018deep}
H.~Nguyen, L.-M. Kieu, T.~Wen, and C.~Cai, ``Deep learning methods in
  transportation domain: a review,'' \emph{IET Intelligent Transport Systems},
  vol.~12, pp. 998--1004, 2018.

\bibitem{wang2018enhancing}
Y.~Wang, D.~Zhang, Y.~Liu, B.~Dai, and L.~H. Lee, ``Enhancing transportation
  systems via deep learning: a survey,'' \emph{Transportation Research Part C:
  Emerging Technologies}, 2018.

\bibitem{petnehazi2019recurrent}
G.~Petneh{\'a}zi, ``Recurrent neural networks for time series forecasting,''
  \emph{arXiv preprint arXiv:1901.00069}, 2019.



\bibitem{fu2016using}
R.~Fu, Z.~Zhang, and L.~Li, ``Using {LSTM} and {GRU} neural network methods for
  traffic flow prediction,'' in \emph{Proc. of the Youth Academic Annual Conference of Chinese Association of Automation,} pp.~324--328, 2016.

\bibitem{chiu2018state}
C.-C. Chiu, T.~N. Sainath, Y.~Wu, R.~Prabhavalkar, P.~Nguyen, Z.~Chen,
  A.~Kannan, R.~J. Weiss, K.~Rao, E.~Gonina, and N.~Jaitly, ``State-of-the-art
  speech recognition with sequence-to-sequence models,'' in \emph{Proc. of the IEEE
  International Conference on Acoustics, Speech and Signal Processing}, pp. 4774--4778, 2018.

\bibitem{yogatama2017generative}
D.~Yogatama, C.~Dyer, W.~Ling, and P.~Blunsom, ``Generative and discriminative
  text classification with recurrent neural networks,'' \emph{arXiv preprint
  arXiv:1703.01898}, 2017.

\bibitem{rutagemwa2018dynamic}
H.~Rutagemwa, A.~Ghasemi, and S.~Liu, ``Dynamic spectrum assignment for land
  mobile radio with deep recurrent neural networks,'' in \emph{Proc. of the IEEE
  International Conference on Communications Workshops,} pp.~1--6, 2018.

\bibitem{yao2018deep}
H.~Yao, F.~Wu, J.~Ke, X.~Tang, Y.~Jia, S.~Lu, P.~Gong, J.~Ye, and Z.~Li, ``Deep
  multi-view spatial-temporal network for taxi demand prediction,'' in
  \emph{AAAI Conference on Artificial Intelligence}, 2018.

\bibitem{liao2018large}
S.~Liao, L.~Zhou, X.~Di, B.~Yuan, and J.~Xiong, ``Large-scale short-term urban
  taxi demand forecasting using deep learning,'' in \emph{Proc. of the Asia and South Pacific Design Automation Conference}, pp.~428--433, 2018.

\bibitem{wang2017deepsd}
D.~Wang, W.~Cao, J.~Li, and J.~Ye, ``Deepsd: supply-demand prediction for
  online car-hailing services using deep neural networks,'' in \emph{Proc. of the IEEE International Conference on Data Engineering,} pp. 243--254, 2017.

\bibitem{wang2018deepstcl}
D.~Wang, Y.~Yang, and S.~Ning, ``Deepstcl: A deep spatio-temporal convlstm for
  travel demand prediction,'' in \emph{Proc. of the International Joint Conference on
  Neural Networks, }pp. 1--8, 2018.

\bibitem{wu2019comprehensive}
Z.~Wu, S.~Pan, F.~Chen, G.~Long, C.~Zhang, and P.~S. Yu, ``A comprehensive
  survey on graph neural networks,'' \emph{arXiv preprint arXiv:1901.00596},
  2019.



\bibitem{li2018diffusion}
Y.~Li, R.~Yu, C.~Shahabi, and Y.~Liu, ``Diffusion convolutional recurrent
  neural network: Data-driven traffic forecasting,'' in \emph{International
Conference on Learning Representations}, 2018.

\bibitem{cui2018high}
Z.~Cui, K.~Henrickson, R.~Ke, and Y.~Wang, ``High-order graph convolutional
  recurrent neural network: A deep learning framework for network-scale traffic
  learning and forecasting,'' \emph{arXiv preprint arXiv:1802.07007}, 2018.

\bibitem{wang2018dynamic}
M.~Wang, B.~Lai, Z.~Jin, X.~Gong, J.~Huang, and X.~Hua, ``Dynamic
  spatio-temporal graph-based {CNN}s for traffic prediction,'' \emph{arXiv
  preprint arXiv:1812.02019}, 2018.

\bibitem{geng2019spatiotemporal}
X.~Geng, Y.~Li, L.~Wang, L.~Zhang, Q.~Yang, J.~Ye, and Y.~Liu, ``Spatiotemporal
  multi-graph convolution network for ride-hailing demand forecasting,'' in
  \emph{AAAI Conference on Artificial Intelligence}, 2019.

\bibitem{davis2018btaxi}
N.~Davis, G.~Raina, and K.~Jagannathan, ``Taxi demand-supply forecasting:
  Impact of spatial partitioning on the performance of neural networks,'' in \emph{NIPS Workshop on Machine Learning in Intelligent Transport Systems}, 2018.

\bibitem{raj2017aggregating}
V.~Raj and S.~Kalyani, ``An aggregating strategy for shifting experts in
  discrete sequence prediction,'' \emph{arXiv preprint arXiv:1708.01744}, 2017.

\bibitem{nyc}
N.~Y. C. T.~L. Comission, ``Tlc trip record data,'' 2018. [Online]. Available:
  http://www.nyc.gov/html/tlc/html/about/trip\_record\_data.shtml.

\bibitem{macqueen1967some}
J.~MacQueen, ``Some methods for classification and analysis of multivariate
  observations,'' in \emph{Proc. of the Berkeley Symposium on
  Mathematical Statistics and Probability}, pp. 281--297, 1967.

\bibitem{chaudhari2012comparative}
B.~Chaudhari and M.~Parikh, ``A comparative study of clustering algorithms
  using {WEKA} tools,'' \emph{International Journal of Application or Innovation
  in Engineering \& Management}, vol.~1, pp.~154--158, 2012.

\bibitem{gers2000learning}
 F. A.~Gers, J.~Schmidhuber, and F.~Cummins, \say{Learning to forget: Continual prediction with LSTM,} \emph{Neural Computation}, vol.~12, pp.~2451--2471, 2000.

\bibitem{xingjian2015convolutional}
S.~Xingjian, Z.~Chen, H.~Wang, D.-Y. Yeung, W.-K. Wong, and W.-C. Woo,
  ``Convolutional {LSTM} network: A machine learning approach for precipitation
  nowcasting,'' in \emph{Advances in Neural Information Processing Systems},  pp. 802--810,
  2015.

\bibitem{shahriari2016taking}
B.~Shahriari, K.~Swersky, Z.~Wang, R.~P. Adams, and N.~De~Freitas, ``Taking the
  human out of the loop: A review of bayesian optimization,'' \emph{Proceedings
  of the IEEE}, vol. 104, pp. 148--175, 2016.

\bibitem{bergstra2011algorithms}
J.~S. Bergstra, R.~Bardenet, Y.~Bengio, and B.~K{\'e}gl, ``Algorithms for
  hyper-parameter optimization,'' in \emph{Advances in Neural Information
  Processing Systems}, pp. 2546--2554, 2011.

\bibitem{freund1995desicion}
Y.~Freund and R.~E. Schapire, ``A decision-theoretic generalization of on-line
  learning and an application to boosting,'' in \emph{Proc. of the European Conference on
  Computational Learning Theory}, pp. 23--37, 1995.

\end{thebibliography}

\begin{IEEEbiography}[{\includegraphics[width=1in,height=1.2in]{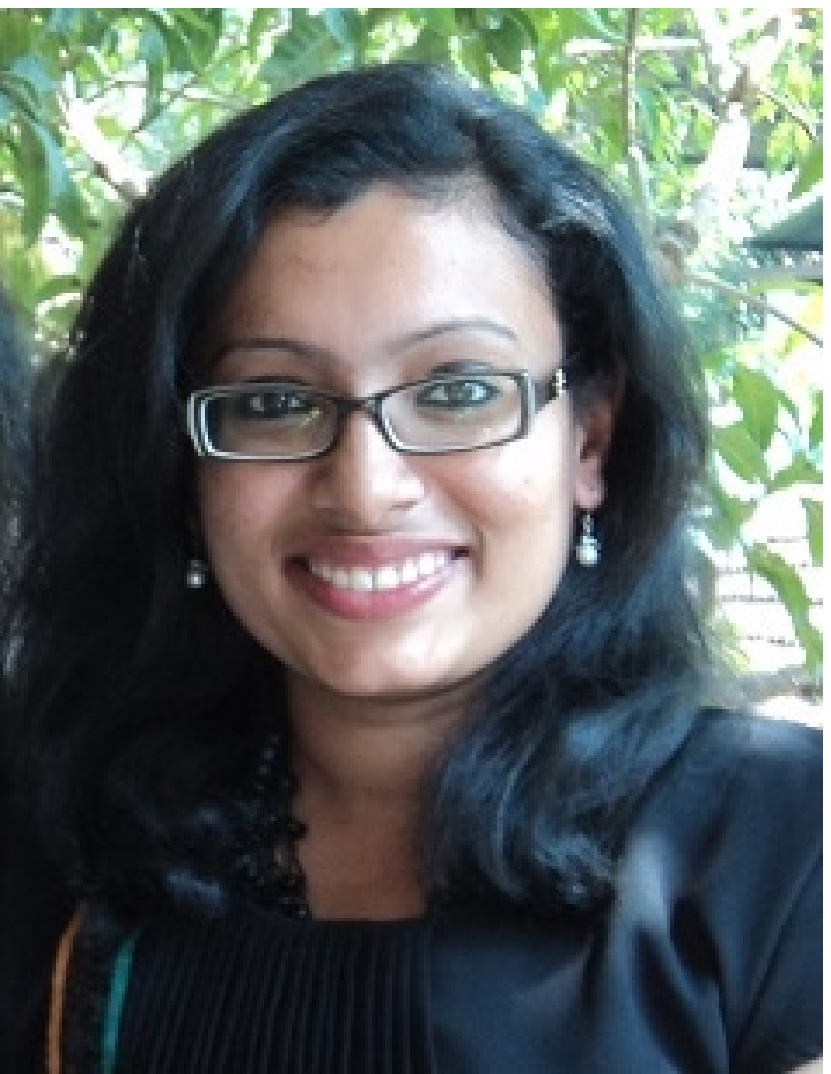}}]{Neema Davis} is currently pursuing her Dual M.S-Ph.D. degree in Electrical Engineering from the Indian Institute of Technology Madras. She obtained her B. Tech. in Electronics and Communication Engineering from the Mahatma Gandhi University Kottayam in 2013. Her research interests lie in transportation planning, predictive data analysis, and intelligent transportation systems.
\end{IEEEbiography}
\vspace{-1cm}
\begin{IEEEbiography}[{\includegraphics[trim={1cm 1cm 10cm 1cm},clip,width=1in,height=1.2in]{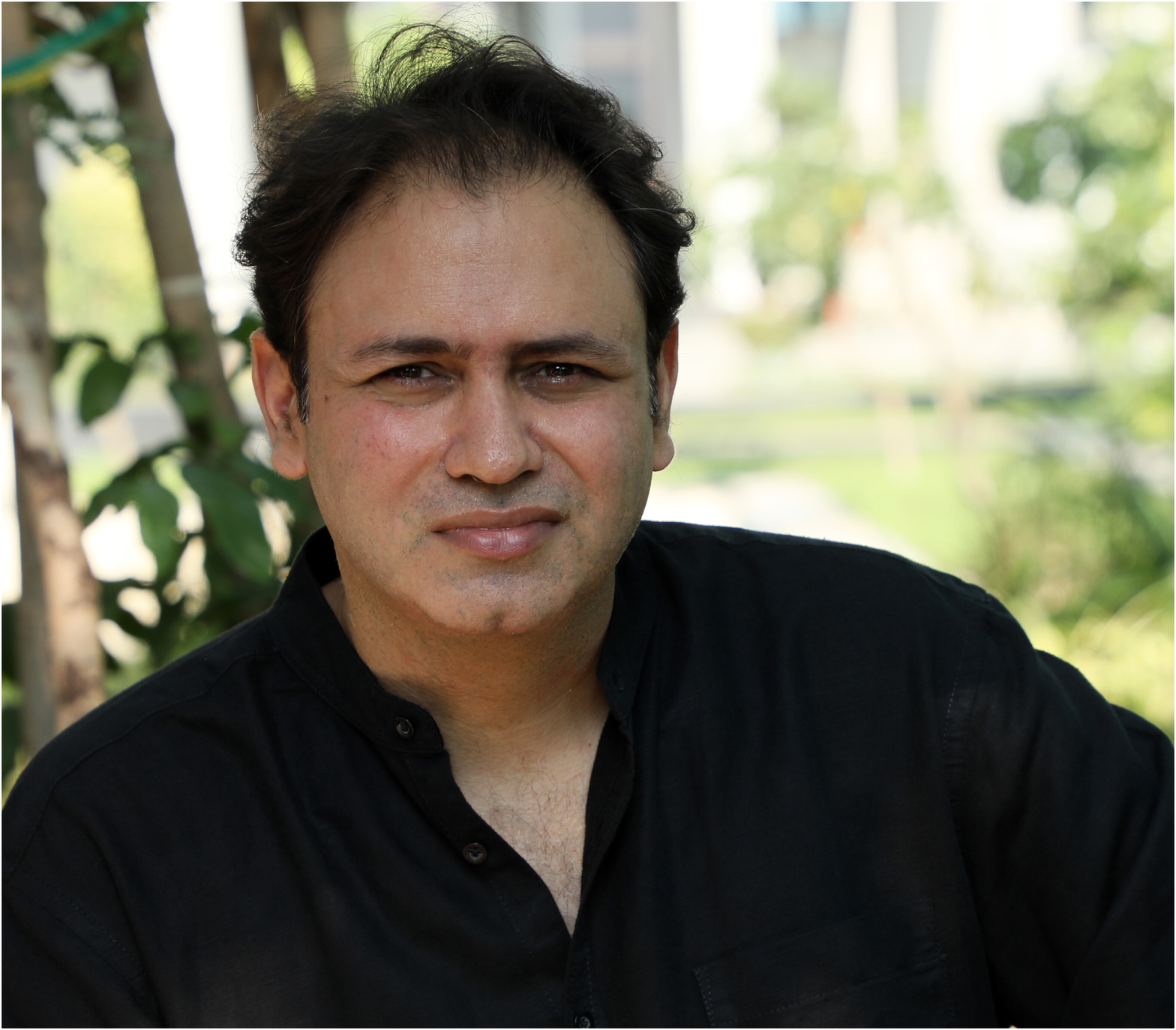}}]{Gaurav Raina} holds a PhD in Mathematics from Trinity College, University of Cambridge, UK. He is a faculty member in Electrical Engineering at IIT Madras, and also a visiting professor of Mathematics and Computer Science at Krea University. He also serves on the Academic Council of Krea University, and is currently the Chairman of the Mobile Payment Forum of India. His research interests span the design and control of large scale engineering systems, like the Internet and intelligent transportation systems.
\end{IEEEbiography}
\vspace{-1cm}

\begin{IEEEbiography}[{\includegraphics[width=1in,height=1.2in]{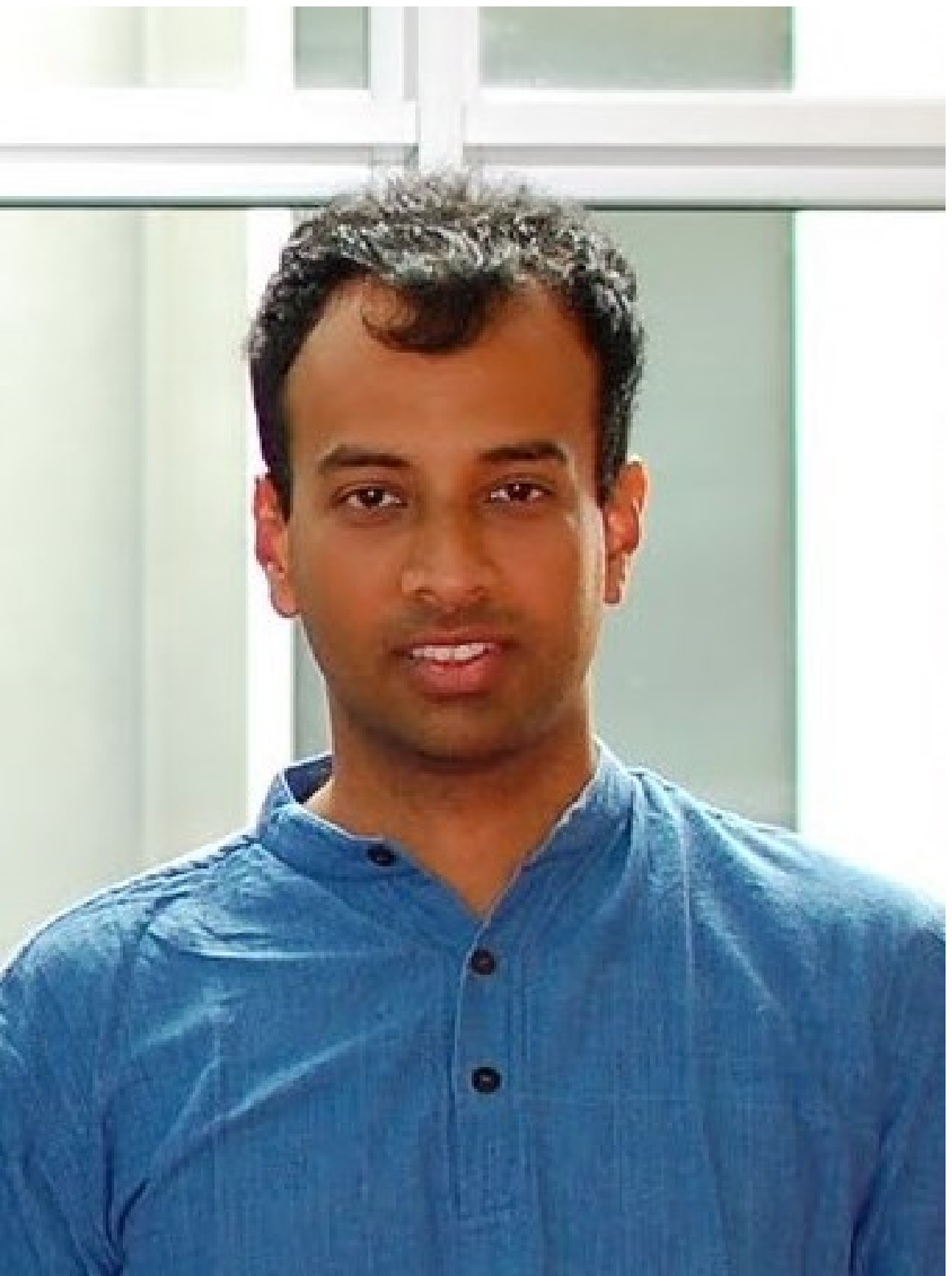}}]{Krishna Jagannathan} obtained his B. Tech. in Electrical Engineering from IIT Madras, and the S.M. and Ph.D. degrees in Electrical Engineering and Computer Science from the Massachusetts Institute of Technology (MIT). He is an associate professor in the Department of Electrical Engineering, IIT Madras. His research interests lie in the stochastic modeling and analysis of communication networks, transportation networks, network control, and queuing theory.  Dr. Jagannathan serves on the editorial boards of the journals \emph{IEEE/ACM Transactions on Networking} and \emph{Performance Evaluation}. 

\end{IEEEbiography}
\end{document}